\documentclass{article} 
\usepackage{colm2024_conference}

\usepackage{booktabs}
\usepackage{graphicx}
\usepackage{enumitem}
\usepackage{wrapfig}
\usepackage{algorithm}
\usepackage{algpseudocode}
\usepackage{natbib}
\usepackage{makecell}
\usepackage{booktabs}
\usepackage{array}
\usepackage{amsmath}
\usepackage{amssymb}
\usepackage{amsfonts}
\usepackage{multirow}
\usepackage{verbatim}
\usepackage{caption}
\usepackage{longtable}
\usepackage{supertabular}
\usepackage{CJKutf8}
\usepackage[utf8]{inputenc} 
\usepackage[T1]{fontenc}
\usepackage[french,vietnamese,mongolian,greek,english]{babel}
\usepackage{pifont}

\usepackage{enumitem}
\usepackage{tablefootnote}
\usepackage{xspace}
\usepackage{textcomp}
\usepackage{makecell}
\usepackage{lscape} 
\usepackage{siunitx}
\usepackage{listings}
\usepackage{xcolor}

\definecolor{dt}{gray}{0.7}
\usepackage[table]{xcolor} 
\usepackage{booktabs}      
\usepackage{graphicx}      
\usepackage{geometry}      
\newcommand{\resultone}[1]{\cellcolor{green!25}#1}
\newcommand{\resulttwo}[1]{\cellcolor{blue!15}#1}
\newcommand{\resultthird}[1]{\cellcolor{yellow!25}#1}
\usepackage{pifont}       
\usepackage{bbding}       
\usepackage{fontawesome}

\usepackage{scrextend}

\usepackage{tgpagella}
\usepackage{latexsym}
\usepackage[T1]{fontenc}
\usepackage[utf8]{inputenc}
\usepackage{microtype}
\definecolor{mydarkblue}{rgb}{0,0.08,0.45}
\definecolor{citecolor}{HTML}{0071BC}
\usepackage{url}            
\usepackage{nicefrac}       
\usepackage{changepage}
\usepackage{xargs}          
\usepackage{wrapfig,lipsum,booktabs}
\usepackage{longtable}
\usepackage{subcaption}
\usepackage{endnotes}

\usepackage{pgfplots}
\usetikzlibrary{pgfplots.groupplots}
\pgfplotsset{compat=1.3}
\usepackage{tikz}
\usetikzlibrary{patterns}

\usepackage[most]{tcolorbox}

\usepackage[capitalize,noabbrev]{cleveref}
\crefname{section}{Section}{\S\S}
\Crefname{section}{Section}{\S\S}
\crefname{table}{Table}{Tables}
\crefname{figure}{Figure}{Figures}
\crefname{algorithm}{Algorithm}{}
\crefname{equation}{eq.}{}
\crefname{appendix}{Appendix}{}
\crefformat{section}{Section #2#1#3}
\usepackage{multicol}
\usepackage{tcolorbox}
\usepackage{titlesec}
\titleformat*{\section}{\large\bfseries}

\title{Dolphin Technical Report}

\author{
\bf Dolphin AI}

\newcommand{\weng}[1]{{\color{black} #1}}

\newcommand{\hc}[1]{{\color{blue} #1}}

\begin{document}

\maketitle

\begin{abstract}
\weng{
Ultrasound is vital in modern medicine but faces challenges like operator dependence, image noise, and dynamic real-time scanning, which hinder effective AI integration. Though large multimodal models excel in other medical imaging areas, they struggle with ultrasound's unique complexities. To bridge this gap, we introduce Dolphin V1 and its reasoning-augmented version, Dolphin R1—
the first multimodal large language models (MLLMs) dedicated to ultrasound, designed to unify diverse clinical tasks within a single vision-language framework.
We curate a 
multimodal dataset at the 2-million-scale, integrating textbook knowledge, diagnostic ultrasound data, synthetic knowledge-distilled samples, and general multimodal corpora. This comprehensive dataset ensures the models achieve robust perception, strong generalization, and broad clinical adaptability across medical domains.}

The Dolphin series adopts a progressive three-stage training strategy that combines domain-specialized pretraining, instruction-driven alignment, and reinforcement-based refinement. The baseline Dolphin V1 establishes reliable performance in classification, detection, regression, and text generation, offering a strong general-purpose model for clinical ultrasound analysis. Building upon this foundation, Dolphin R1 leverages reinforcement learning with verifiable ultrasound-specific reward signals, thereby enabling deeper diagnostic inference, enhanced reasoning transparency, and more interpretable decision pathways.

To evaluate the models, we employ U2-Bench for ultrasound understanding and additionally assess performance on other medical benchmarks and general benchmarks. Experimental results demonstrate that Dolphin R1 achieves a U2-Score of 0.5835, more than twice the score of the second-best model (0.2968), establishing a new state of the art in multimodal ultrasound understanding. On broader benchmarks, Dolphin also achieves competitive or superior performance, with strong gains in general perception abilities (MME) and medical temporal reasoning (MedFrameQA). 
Further comparison between standard mode (V1) and deep reasoning mode (R1) reveals that reasoning-enhanced training significantly boosts diagnostic accuracy, consistency, and interpretability. These findings underscore the importance of integrating reasoning into MLLMs for medical domains. Together, we provide a scalable paradigm for future developments in intelligent ultrasound analysis.
\end{abstract}
\begin{figure*}[ht]
\centering
\includegraphics[width= 1\linewidth]{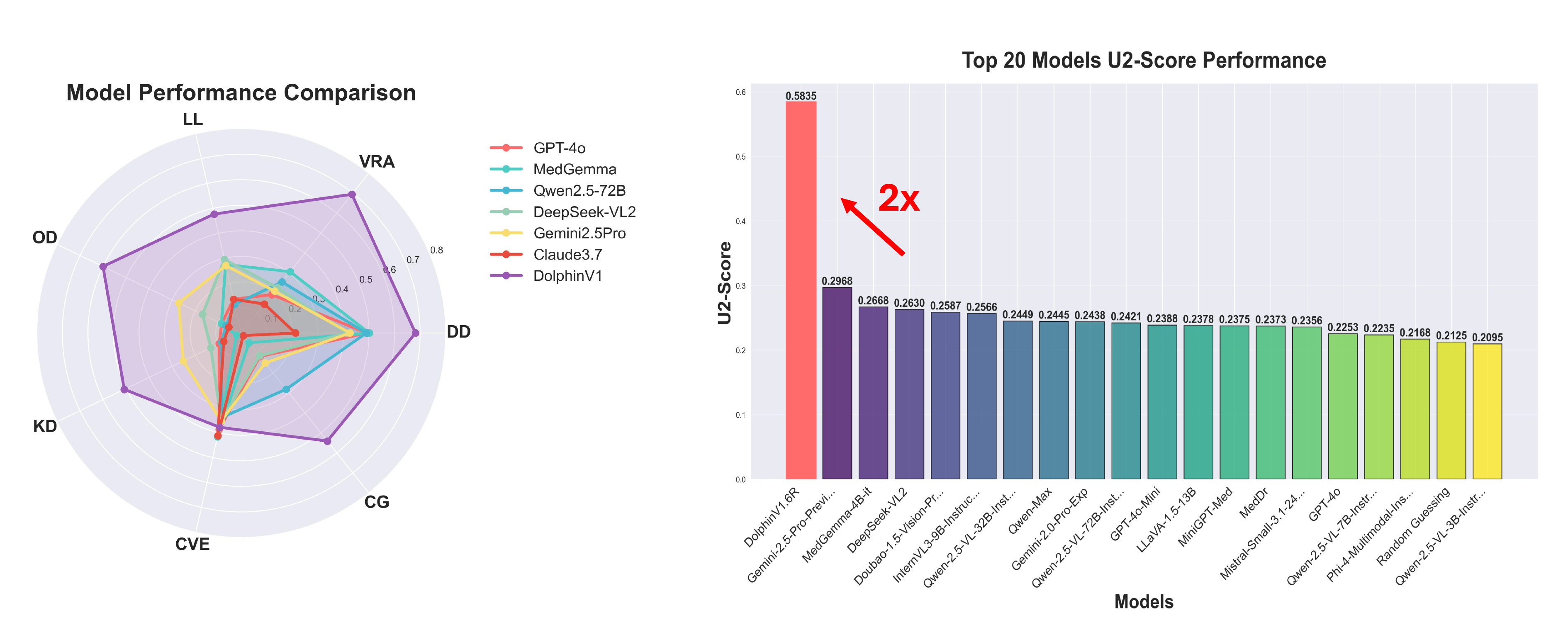}
\end{figure*}

\newpage

\section{Introduction}


Currently, multimodal large language models (MLLMs) have demonstrated significant potential in the medical field, offering advanced capabilities for integrating and analyzing diverse data types, including medical images (MRI,CT,PET; \cite{lee2024read,volkov2025visual,shui2025large}), electronic health records (\cite{alsaad2024multimodal}), genomic data (\cite{fallahpour2025bioreason}), and clinical notes (\cite{han2024ascleai}). While these models have achieved remarkable success in processing static, high-resolution medical images and structured clinical data, their application to ultrasound imaging presents unique challenges that remain largely unaddressed.

Ultrasound is widely used in clinical practice due to its real-time capability, portability, safety, and cost-effectiveness \citep{hewson2023closing}. However, the translation of multimodal approaches to ultrasound is hindered by several critical limitations. First, unlike the standardized acquisition protocols for CT or MRI, ultrasound examinations are highly operator-dependent, resulting in significant variability in image quality and anatomical views (\cite{stasi2015critical}). Second, the dynamic nature of ultrasound video sequences - often containing crucial temporal information about tissue motion and blood flow - is fundamentally different from the static images that current MLLMs typically process (\cite{szabo2013diagnostic}). Third, the characteristic speckle noise and lower signal-to-noise ratio of ultrasound images compared to other modalities create additional challenges for feature extraction and interpretation (\cite{loizou2022introduction}). Furthermore, existing MLLMs struggle with the sparse and often subjective nature of ultrasound reports, which typically lack the detailed anatomical descriptions found in radiology reports for other imaging modalities (\cite{wang2024survey}). These challenges highlight the need for MLLMs capable of processing the unique characteristics of ultrasound imaging.


Medical ultrasound understanding requires precise perception of visual features, along with sophisticated reasoning to interpret dynamic, operator-dependent imaging and derive clinically meaningful conclusions. However, most existing multimodal medical models treat ultrasound tasks in isolation and focus primarily on perceptual accuracy, neglecting the development of systematic reasoning capabilities. Moreover, the scarcity of real-world diagnostic reasoning data in clinical settings poses a significant barrier to training models with robust medical reasoning skills. Recognizing these challenges, we introduce Dolphin, a general-purpose multimodal ultrasound understanding model capable of jointly interpreting ultrasound images, videos, dynamic scanning sequences, and associated clinical text. By leveraging a three-stage training framework with hybrid reasoning, Dolphin achieves deep cognitive capabilities in ultrasound analysis. 

To support model development, we curate a large-scale multimodal ultrasound dataset containing over 2 million high-quality samples sourced from medical textbooks, public repositories, clinical guidelines, and synthetic data. Additionally, we propose the Dolphin Ultrasound Data Protocol (DUDP), a standardized data interface that unifies and formalizes a wide range of ultrasound tasks, unifying ultrasound tasks into four categories: classification, detection, regression, and generation.

 To effectively train a medical ultrasound language model with strong generalization and diagnostic reasoning capabilities, we adopt a progressive three-stage training strategy. The first stage leverages post-training on mixture of ultrasound-specific and general-domain data to build  both medical foundation and broad language understanding.  The second stage focuses on instruction tuning with multi-turn dialogues and deep-reasoning tasks, enhancing the model’s ability to engage in complex clinical discussions and perform step-by-step analysis. The final stage introduces UARPO, a reinforcement learning method guided by Ultrasound Answer Reward (UAR), which combines assessment of formatted answers and alignment with true diagnostic outcomes.

Dolphin achieves state-of-the-art performance with a U2-Bench~\citep{le2025u2bench} score of 0.5835, outperforming all compared methods across eight ultrasound tasks. During the first stage of post-training, we train the model using easily accessible ultrasound QA data along with pure-text deep reasoning datasets from other domains. We observe that the model spontaneously developed a deep reasoning capability in ultrasound vision. We unify vanilla and deep reasoning within a Bayesian framework by modeling the latent chain-of-thought variable $c$. By introducing mode-dependent priors $P_\theta(c \mid x, m)$, we theoretically characterize cross-domain reasoning transfer: although deep reasoning data lacks target-domain samples, structured reasoning patterns generalize via shared parameters and prior bias. Our results highlight a critical threshold in reasoning data volume: effective reasoning emerges only when training exceeds this point. Meanwhile, experimental results show that deep reasoning mode improves diagnostic accuracy by 2.4\%, reduces measurement RMSE by 10.6\%, and increases detection accuracy by 16\%. We find that base model performance determines the UARPO phase ceiling, with stronger foundation models achieving higher reinforcement learning gains.

Our key contributions can be summarized as follows:

\begin{itemize}
\item We introduce Dolphin, the first multimodal large language model tailored for ultrasound reasoning, with chat and reasoning versions. We also build a large-scale ultrasound dataset of over 2 million samples covering 16 body regions to enhance the model’s foundational capabilities.
\item We adopt a three-stage training strategy with mixed reasoning and UARPO-based reinforcement learning, using deep-reasoning data to improve the model’s visual reasoning for ultrasound. Ultimately, Dolphin successfully develops an emergent capacity for deep reasoning about ultrasound.
\item Dolphin achieves state-of-the-art performance on U2-Bench, surpassing all existing models across all six tasks with a remarkable U2-Score of 0.58, while also demonstrating strong generalization capabilities on mainstream benchmark datasets. Notably, Dolphin has been successfully deployed on real-world ultrasound devices, marking a significant step toward practical clinical integration by bringing AI directly into physicians' daily workflows.
\end{itemize}

\section{Data Curation}

\subsection{Post-training data}
\begin{figure*}[t]
\centering
\includegraphics[width= 1\linewidth]{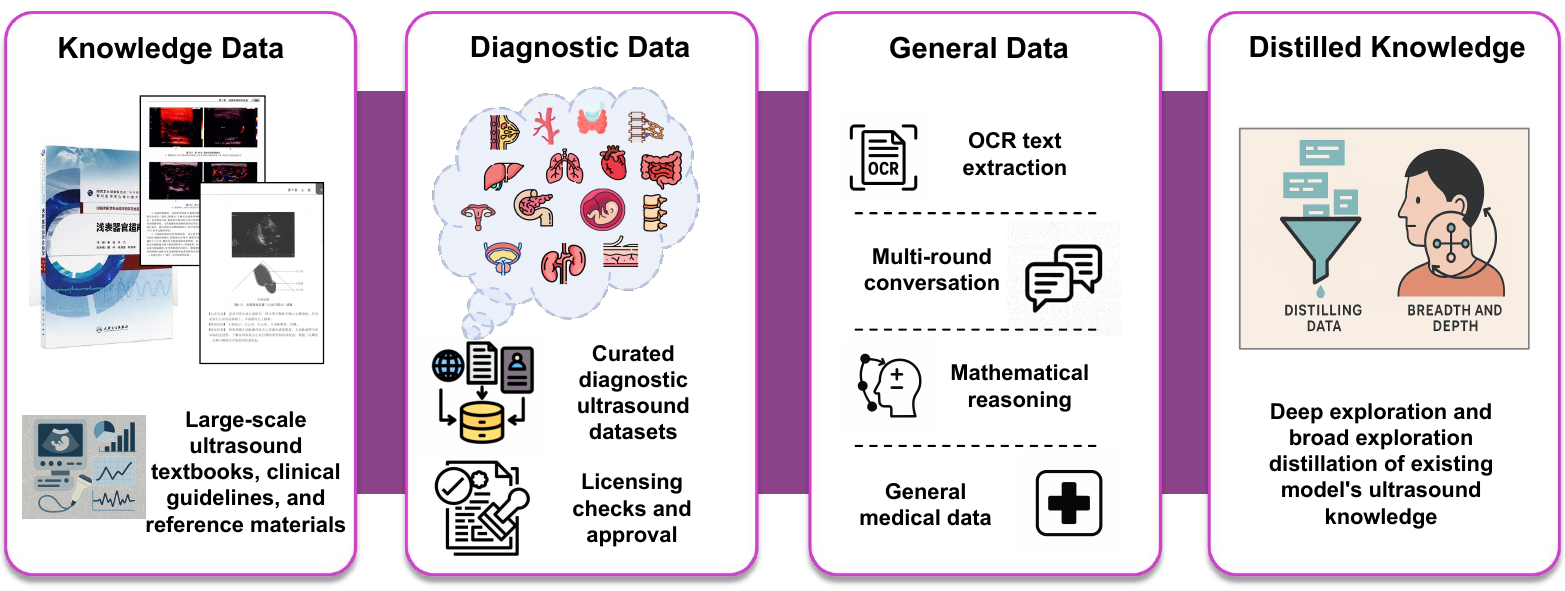}
\caption{Data types of the Dolphin Ultrasound Large-Scale Dataset}
\label{fig:data_types}
\end{figure*}

To enable the model to possess expert-level ultrasound image recognition capabilities, we constructed a large-scale, high-quality ultrasound multimodal dataset. Specifically, our data is primarily covering the following four aspects: (1) \textbf{Ultrasound Knowledge-Oriented Data} - to enhance the model's fundamental knowledge about ultrasound; (2) \textbf{Ultrasound Diagnostic-Oriented Data} - covering classification, detection, regression and other tasks, to improve the model's disease diagnostic abilities; 
(3) \textbf{Ultrasound Domain Distillation Data} - to strengthen the model's instruction-following and formatted output abilities; (4) \textbf{General Purpose Data} - to enhance the model's general and conversational capabilities. An illustration is included in Figure \ref{fig:data_types}.


Our data generation pipeline was meticulously designed to minimize the risk of hallucination and ensure high fidelity in clinical representation. The process began with a well-defined task framework and template construction, which was followed by an initial review by board-certified physicians to validate the clinical relevance and diagnostic coherence of the proposed structure. Upon achieving expert consensus, synthetic data was generated in accordance with the predefined templates. This synthesized data then underwent a second round of rigorous evaluation by medical professionals, who provided detailed feedback to guide iterative refinement and correction. 
After achieving substantial alignment with clinical standards, the data advanced to a final validation stage, where it was independently reviewed by physicians before inclusion in the training corpus. 
In total, this systematic approach incorporated three iterative rounds of cross-validation, ensuring robust clinical plausibility, diagnostic accuracy, and methodological rigor throughout the data creation process.

\subsubsection{Ultrasound Knowledge-Oriented Data}

We collected a large number of classic ultrasound textbooks and industry guidelines, uniformly converted these materials into Markdown format, manually filtered out non-substantive content such as covers, back covers, and margins, and corrected grammatical errors and formulas that were inaccurately converted. Based on these ultrasound textbooks, we constructed the following three datasets: the \textbf{Ultrasound Question Answering}, the \textbf{Ultrasound Caption}, and the \textbf{Ultrasound Image-Text Explanation}.

\textbf{Ultrasound Question Answering.} We segmented the textbook data into chunks and employed GPT to generate contextually appropriate questions for each segment. Through self-distillation, we regenerated corresponding answers to ensure strict question-answer consistency.

\textbf{Ultrasound Caption.} The ultrasound textbook corpus contains systematic image-caption pairs, which we processed into a structured dataset through automated format matching to ensure precise image-annotation correspondence. It should be noted that our constructed caption dataset contains cases with multiple images, which helps enhance the model's multi-image comprehension capability.

\textbf{Ultrasound Image-Text Explanation.} The ultrasound textbooks contained detailed textual descriptions of medical images. We first identified the text passages corresponding to each image, then processed these descriptions through GPT using self-distillation to generate high-quality question-answer pairs. This approach ensures data accuracy while simultaneously enhancing the model's response diversity and instruction-following capabilities.

\subsubsection{Ultrasound Diagnostic-Oriented Data}

To enhance the model's diagnostic capabilities across various tasks, we collected a substantial number of diagnostic-oriented ultrasound datasets and established a unified data processing pipeline. We developed a unified file standard called \textbf{Dolphin Ultrasound Data Protocol (DUDP)} to harmonize diverse ultrasound tasks by converting them into a standardized JSON format. The protocol comprehensively handles diverse ultrasound tasks, supporting multimodal data types including ultrasound images, video sequences, segmentation masks, and associated quantitative measurements. 

\begin{figure*}[t]
\centering
\includegraphics[width= 1\linewidth]{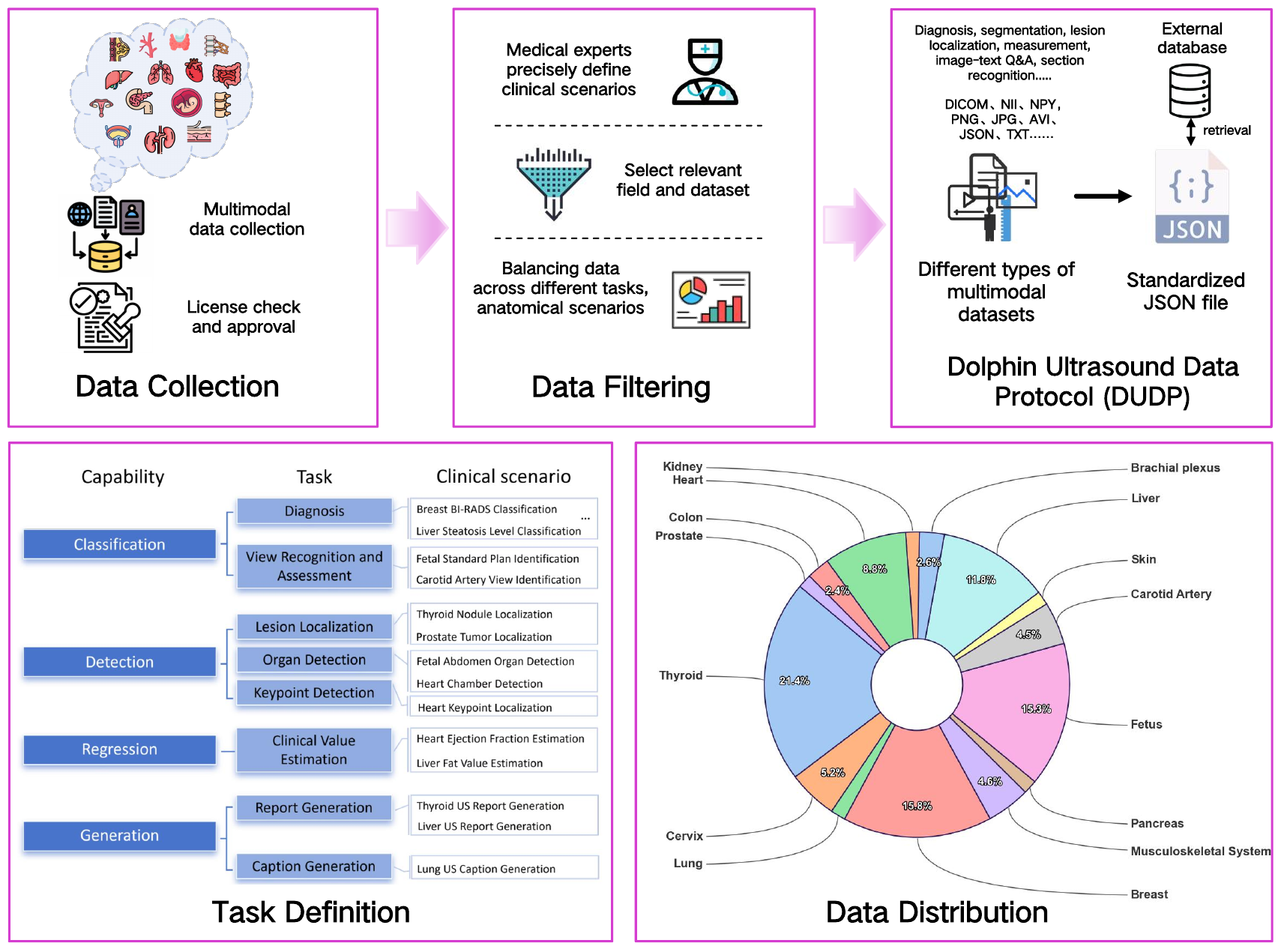}
   \caption{Overview of the Dolphin Ultrasound Diagnostic Data Pipeline. The framework includes data collection, expert-guided filtering, and standardization via the Dolphin Ultrasound Data Protocol (DUDP), ensuring consistency across multimodal sources. Tasks are unified into four categories—classification, detection, regression, and generation—covering diverse clinical scenarios. The dataset distribution demonstrates balanced coverage across major anatomical regions, supporting robust model generalization.
}
\label{fig:public}
\end{figure*}

Specifically, we categorized ultrasound tasks into the following standardized classes: (1) \textbf{Anatomical Recognition}
(2) \textbf{Standard View Identification} (3) \textbf{Diagnostic Classification}
(4) \textbf{Tissue Segmentation}
(5) \textbf{Biometric Measurement}.  Corresponding ultrasound question-answering datasets were constructed for each task category. To mitigate hallucination risks, we developed over a hundred fixed question templates for each task category. For each sample, we randomly selected a question template and populated it with JSON-formatted content, thereby constructing a high-quality ultrasound question-answering dataset. Figure \ref{fig:public} presents an authentic diagnostic data sample from our constructed dataset.

Furthermore, to ensure data diversity, we employed GPT for partial data augmentation and leveraged an LLM-based rating scheme to evaluate response quality, guaranteeing that the JSON-formatted content remains embedded within the generated responses.

\subsubsection{Ultrasound Domain Distillation Data}

Following the methodology of \citet{xu2023wizardlm}, we first constructed a set of ultrasound data question templates. The difficulty level of questions was enhanced through depth-wise augmentation, while their diversity was improved via breadth-wise expansion. All question categories were strictly standardized within the ultrasound medical domain, and we distilled ultrasound-specific data from both GPT and DeepSeek. Additionally, we designed a data screening pipeline that employs large language models to assess the logical consistency of outputs, ultimately yielding a high-quality distilled dataset.

\begin{figure*}[ht]
\centering
\includegraphics[width= 1\linewidth]{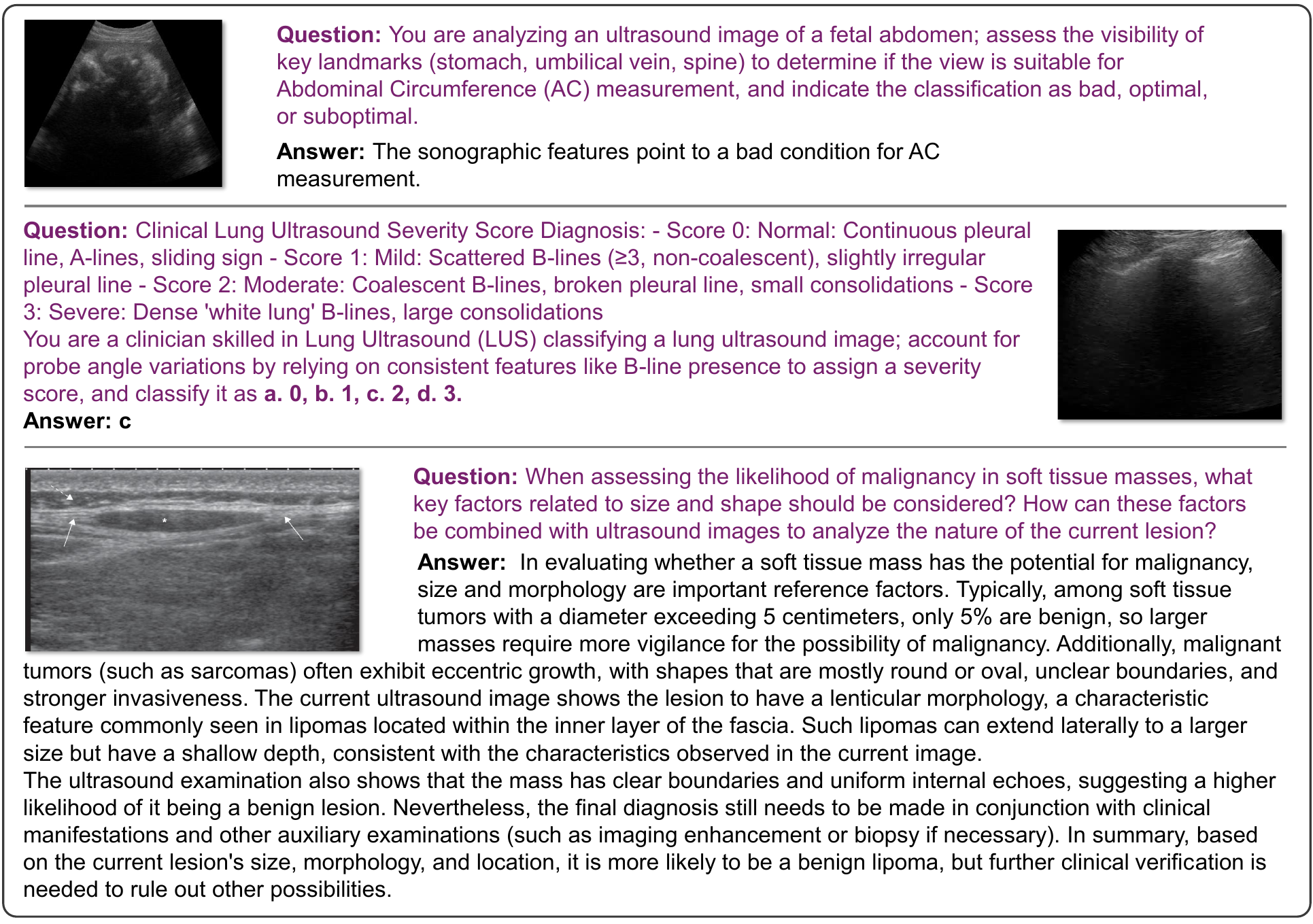}
\caption{Examples of constructed ultrasound question–answer pairs. The dataset integrates diverse clinical tasks, including fetal abdominal view assessment, lung ultrasound severity scoring, and soft tissue lesion evaluation, with corresponding expert-verified answers. These QA pairs provide structured supervision signals that enhance both low-level image understanding and high-level diagnostic reasoning in Dolphin models.}
\end{figure*}

\subsubsection{General Purpose Data}

To enhance the model's general capabilities while incorporating ultrasound-specific knowledge, we selected corresponding general-ability datasets based on different competency requirements. The included datasets were primarily utilized to enhance the following model capabilities: (1) \textbf{Comprehension}; (2)  \textbf{Conversational}; (3) \textbf{Deep reasoning}; (4) \textbf{Medical}; (5) \textbf{Image-text dialogue}.

\subsection{Data Filtering}

To ensure the quality and reliability of the dataset, we designed a multi-stage filtering pipeline. In the first stage, \textbf{model-based semantic filtering} was applied, where LLM-based scoring (e.g., GPT-4) was used to evaluate the \emph{relevance} of answers to the questions, their \emph{accuracy} with respect to medical knowledge, and their \emph{logical consistency}. This step helped remove outputs that were factually incorrect, irrelevant, or incoherent. 

Next, we performed \textbf{rule-based filtering}. This included character matching to verify JSON integrity and caption consistency, as well as format screening to eliminate samples with overly short responses, excessive numbers of images, or the presence of sensitive information. 

Finally, we conducted \textbf{expert validation}, in which board-certified physicians reviewed the remaining samples to assess the rationality of dataset construction and excluded any clinically unreasonable cases.

\section{Model Training}

Dolphin V1 is built upon the Qwen2.5-VL model architecture, which is an advanced large vision-language model developed by \citet{bai2025qwen2}. We enhanced the original framework with a redesigned Vision Transformer (ViT) for optimized visual processing.

We systematically trained two model variants: a base version with 7 billion parameters and an extended version with 72 billion parameters. For each scale, we optimized the training configurations separately, with particular emphasis on investigating the impact of training epochs on both the generalizability and domain-specific performance of the models.

\subsection{Training Recipe}

The \textbf{Dolphin} training pipeline follows a three-stage progressive framework, designed to integrate domain-specific knowledge, align with human preferences, and refine autonomous decision-making capabilities. The specific three stages are as follows: (1) \textbf{Domain-Specialized Training}; (2) \textbf{Instruction-Driven Alignment}; (3) \textbf{Autonomous Reinforcement Refinement}. Through two key phases - domain knowledge injection and instruction alignment - we developed the \textbf{Dolphin V1} model. Building upon this foundation, we conducted reinforcement learning training using extensive ultrasound image-text data to obtain the \textbf{Dolphin R1} model. Our subsequent research further explored the significance and impact of reinforcement learning on large-scale medical models.

\begin{figure*}[ht]
\centering
\includegraphics[width= 1\linewidth]{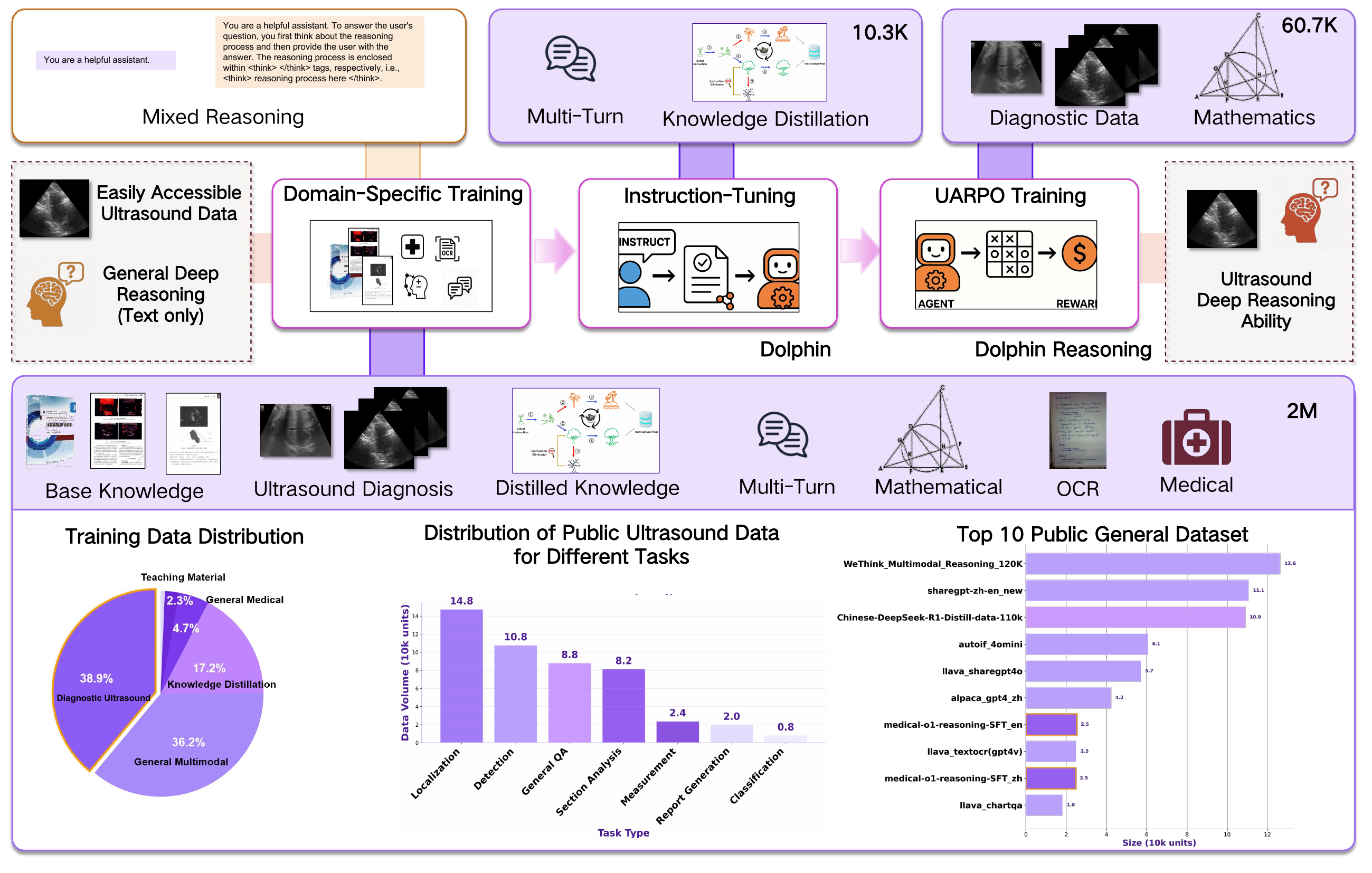}
\caption{Training pipeline of the Dolphin series model. The process consists of three key stages: domain-specific pretraining on 2M multimodal medical data, instruction finetuning with 10.3k curated samples to enhance alignment with clinical tasks, and self-play reinforcement learning on 60.7k interactions to further improve reasoning and decision-making ability.}
\end{figure*}

\subsubsection{Domain-Specialized Training}
The goal of the \textbf{Domain-Specialized Training} is to inject ultrasound-specific knowledge into the large language model while preserving its inherent generalizability to the greatest extent possible. During this phase, the model is trained on extensive textbook-based ultrasound data and diagnostic-oriented ultrasound datasets. The training encompasses ultrasound images of 15 major anatomical systems, with particular emphasis on developing fundamental capabilities in \textbf{disease diagnosis}, \textbf{anatomical localization}, and \textbf{standard plane recognition}. 

Recent studies \citep{chen2020recall,yang2025fine} evidence the occurrence of catastrophic forgetting when fine-tuning LLMs on medical datasets, thereby compromising their core linguistic competencies, highlighting the need for balanced adaptation strategies in medical applications. 
Therefore, when fine-tuning the model with ultrasound data using SFT (Supervised Fine-Tuning), we incorporated data from other domains, with particular emphasis on the model's fundamental capabilities in the following aspects: \textbf{reasoning}, \textbf{multi-turn dialogue}, \textbf{instruction-following }, \textbf{OCR detection}, and \textbf{mathematical reasoning}, which helped to maintain the core competencies of the original LLM. 
A figure illustrating the model's performance on other benchmarks is included in Figure \ref{deep_reasoning}.

\begin{figure*}[t]
\centering

\includegraphics[width= 1\linewidth]{ 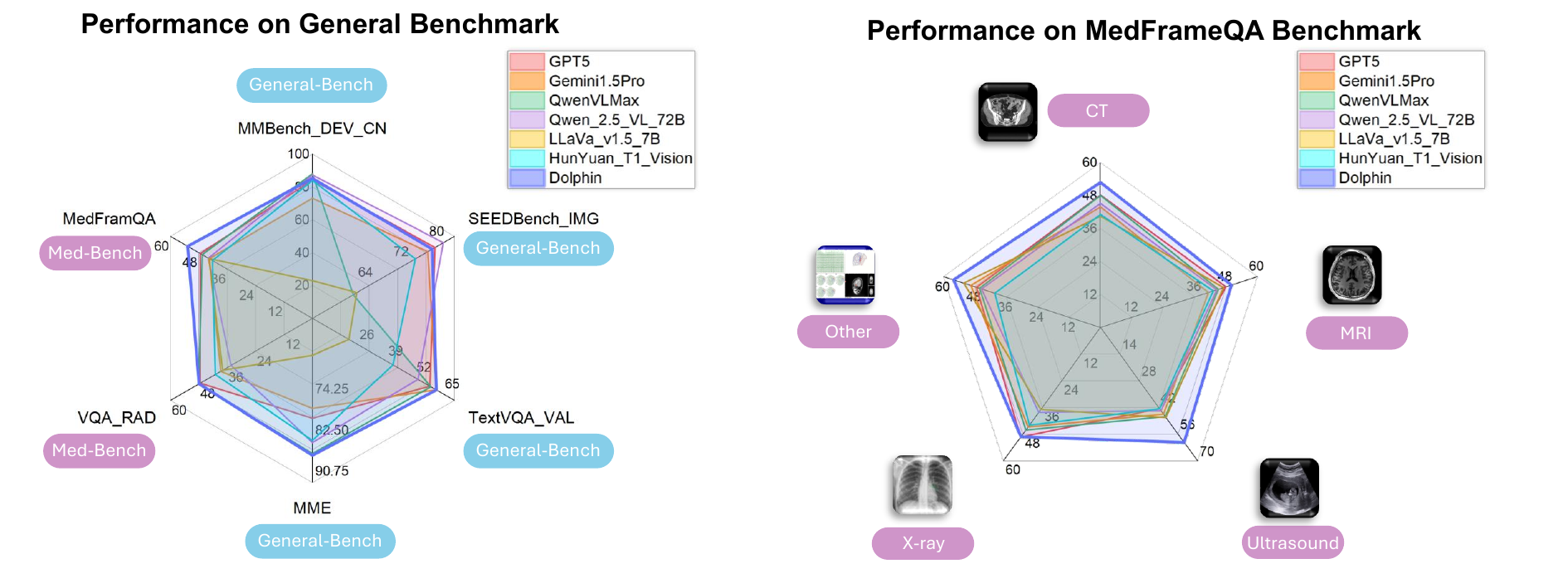}
\caption{Model performance on General \& Medical Bench}
\label{deep_reasoning}
\end{figure*}

Interestingly, we observed that as the volume of both ultrasound-specific and general-domain data increased proportionally, the model demonstrated simultaneous improvements in both ultrasound-related tasks and general-domain performance. This synergistic enhancement suggests that domain-specific specialization and general capability maintenance can be mutually reinforcing when balanced appropriately during training.

\subsubsection{Instruction-Driven Alignment}

During the \textbf{Instruction-Driven Alignment} phase, our primary focus is refining the model’s output to ensure strict adherence to predefined formats and content requirements. This involves fine-tuning the model’s responses to align with structured templates, domain-specific conventions, or user-specified guidelines. Our methodology combines automated techniques with expert human oversight:

\begin{itemize}
    \item \textbf{Distilled Knowledge Integration:}
    \begin{itemize}
        \item We constructed a small-scale instruction dataset primarily derived from distilled knowledge data. This dataset captures high-quality, structured responses from existing aligned models. It enables the model to learn and replicate the desired stylistic patterns, ensuring consistency with established outputs while minimizing bias.
    \end{itemize}
    \item \textbf{Expert-Guided Refinement:} 
    \begin{itemize}
        \item Submit first-phase model outputs to domain specialists (e.g., medical professionals) for review.
        \item Incorporate expert feedback through direct annotation of inaccuracies.
        \item Implement iterative refinement cycles to progressively improve output quality
    \end{itemize}
\end{itemize}

At this stage, to prevent degradation of the model's inherent capabilities, we adopted a relatively small learning rate and performed only a few epoch of fine-tuning on the constructed Instruction-Driven Alignment dataset.

\subsubsection{Autonomous Reinforcement Refinement}

Recent advances in reasoning models, such as OpenAI’s o-series \citep{jaech2024openai} and DeepSeek-R1 \citep{guo2025deepseek}, highlight the importance of reinforcement learning (RL) in enhancing diagnostic reasoning and interpretability. Unlike supervised fine-tuning (SFT), which is prone to overfitting and shortcut learning \citep{hao2025understanding}, RL enables models to autonomously explore reasoning pathways, thereby improving robustness and generalization in high-stakes domains such as medicine \citep{pan2025medvlm}.

For ultrasound-specific reinforcement learning, we designed the \textbf{Ultrasound Answer Reward (UAR)}, which evaluates model responses along two axes: format compliance and diagnostic accuracy. The reward is defined as:
\begin{equation}
\text{UAR}(x, y) = 0.5 \times R_{\text{format}}(y) + 0.5 \times R_{\text{outcome}}(x, y),
\end{equation}
where $R_{\text{format}}(y) = 1$ if the output follows the required structured format, and $R_{\text{outcome}}(x, y) = 1$ if the response matches the ground-truth outcome. The outcome reward encompasses two types of verifiable results: (1) multiple-choice answers (A/B/C/D) for classification-style tasks, and (2) clinically accurate diagnostic conclusions for open-ended tasks, including disease identification, anatomical localization, and quantitative measurements. 

This binary reward system ensures that responses are both structurally consistent and clinically valid. During reinforcement training, we observed that performance improvements strongly depended on the representational capabilities learned in the SFT phase; reinforcement learning primarily activated and refined these capabilities rather than introducing new medical knowledge.

\section{Experiment}
To evaluate the ultrasound understanding capabilities of the trained model, we adopt U2-Bench~\citep{le2025u2bench}, a comprehensive benchmark specifical designed for ultrasound-related tasks. The experimental results presented in Table~\ref{tab:mainresult} provide a comprehensive evaluation of various MLLMs on the U2-Bench, a benchmark specifically designed to assess ultrasound-related multimodal understanding capability. Among the models evaluated, the Dolphin series, particularly Dolphin V1, Dolphin R1, and their smaller variant Dolphin V1-7B, demonstrate competitive performance across multiple tasks, with Dolphin R1 achieving the highest U2-Score of 0.5835 among all models tested.

\begin{table}[t]
\centering
\renewcommand{\arraystretch}{1.2} 
\caption{\textbf{Results of different models on the U2-Bench.} To calculate the \texttt{\textbackslash score{}} for random guessing, the BLEU scores are taken to be zero.}
\label{tab:mainresult}

\resizebox{\textwidth}{!}{%
\setlength{\tabcolsep}{3.0pt}
\begin{tabular}{l|cc|cc|c|c|c|ccc|ccc|ccc|c}
\toprule
\textbf{Models} & \multicolumn{2}{c|}{\textbf{DD}} & \multicolumn{2}{c|}{\textbf{VRA}} & \textbf{LL} & \textbf{OD} & \textbf{KD} & \multicolumn{3}{c|}{\textbf{CVE}} & \multicolumn{3}{c|}{\textbf{RG}} & \multicolumn{3}{c|}{\textbf{CG}} & \textbf{U2-Score} \\
\cline{2-18} 
~ & Acc. & F1 & Acc. & F1 & Acc. & Acc. & Acc. & RMSE & MAE & \%tol & BLEU\% & Rouge\% & BERT\% & BLEU\% & Rouge\% & BERT\% & ~ \\
\midrule
Random Guessing & 0.4143 & 0.4135 & 0.3195 & 0.3184 & 0.1118 & 0.0680 & 0.1120 & 0.5472 & 0.4352 & 18.776 & - & - & - & - & - & - & 0.2125 \\
\midrule
\multicolumn{18}{c}{\textit{Medical-Specific Open-Source Models}} \\
\midrule
MiniGPT-Med & 0.3468 & 0.2828 & 0.1800 & 0.1048 & 0.1728 & 0.1789 & 0.0840 & 0.3056 & 0.2600 & 33.2259 & \resulttwo{6.4700} & \resulttwo{20.1300} & \resulttwo{74.6900} & 30.2000 & 47.7500 & 80.5000 & 0.2375 \\
MedDr & 0.4508 & 0.3118 & 0.2071 & 0.1214 & 0.0720 & 0.0881 & 0.0900 & 0.2144 & 0.1786 & 38.2642 & 2.7998 & 13.5060 & 72.2050 & 33.4939 & 49.6236 & 81.2078 & 0.2373 \\
\midrule
\multicolumn{18}{c}{\textit{Open-Source Multimodal Models}} \\
\midrule
Qwen-2.5-VL-3B-Instruct & 0.4503 & 0.3591 & 0.2097 & 0.1492 & 0.0696 & 0.0649 & 0.0894 & 0.5008 & 0.4519 & 18.9055 & 3.5018 & 15.0327 & 72.8419 & 27.6748 & 44.7618 & 79.8849 & 0.2095 \\
Qwen-2.5-VL-7B-Instruct & 0.4821 & 0.3860 & 0.2181 & 0.1665 & 0.0750 & 0.0704 & 0.1000 & 0.4646 & 0.4337 & 19.7115 & 3.7100 & 15.5600 & 73.1500 & 29.4400 & 47.0000 & 81.1500 & 0.2235 \\
Qwen-2.5-VL-3B-Instruct     & 0.4503 & 0.3591 & 0.2097 & 0.1492 & 0.0696 & 0.0649 & 0.0894 & 0.5008 & 0.4519 & 18.9055 & 3.5018 & 15.0327 & 72.8419 & 27.6748 & 44.7618 & 79.8849 & 0.2095 \\
Qwen-2.5-VL-7B-Instruct     & 0.4821 & 0.3860 & 0.2181 & 0.1665 & 0.0750 & 0.0704 & 0.1000 & 0.4646 & 0.4337 & 19.7115 & 3.7100 & 15.5600 & 73.1500 & 29.4400 & 47.0000 & 81.1500 & 0.2235 \\
Qwen-2.5-VL-32B-Instruct    & 0.4812 & 0.3860 & 0.2864 & 0.2071 & 0.1700 & 0.0755 & 0.0880 & 0.3414 & 0.3015 & 27.4038 & 1.1900 & 13.0100 & 68.1400 & 14.7700 & 38.6800 & 77.3900 & 0.2449 \\
Qwen-2.5-VL-72B-Instruct    & 0.4895 & 0.4556 & 0.2559 & 0.1789 & 0.1150 & 0.0660 & 0.0860 & 0.3224 & 0.2733 & 37.9370 & 3.0900 & 15.0600 & 72.6600 & 28.1600 & 44.2800 & 80.9100 & 0.2421 \\
DeepSeek-VL2                & 0.4126 & 0.3190 & 0.2268 & 0.1111 & 0.2950 & 0.1682 & 0.1320 & 0.2956 & 0.2505 & 12.3355 & \resultone{7.4700} & \resultone{20.5400} & \resultone{75.3800} & 11.4200 & 34.8500 & 77.2400 & 0.2630 \\
InternVL3-9B-Instruct       & 0.4447 & 0.3716 & 0.1926 & 0.1083 & 0.3000 & 0.1416 & 0.0940 & 0.2429 & 0.1733 & 50.8738 & 2.1600 & 14.7000 & 72.2100 & 21.5900 & 43.1300 & 80.9800 & 0.2566 \\
LLaVA-1.5-13B               & 0.4321 & 0.3055 & 0.1731 & 0.0755 & 0.1700 & 0.1259 & 0.1100 & 0.2307 & 0.1976 & 24.7964 & \resultthird{6.2400} & \resultthird{18.5800} & 73.7900 & 10.8300 & 29.4000 & 75.5000 & 0.2378 \\
Phi-4-Multimodal-Instruct   & 0.3686 & 0.1148 & 0.2452 & 0.0537 & 0.0350 & 0.0815 & 0.1600 & 0.2249 & 0.2006 & 16.1972 & 3.2700 & 16.5800 & 73.2700 & 3.8700 & 22.9800 & 73.0800 & 0.2168 \\
Mistral-Small-3.1-24B-Inst. & 0.4359 & 0.0936 & 0.1964 & 0.0664 & 0.1300 & 0.0910 & 0.1060 & \resultthird{0.1675} & \resultthird{0.1331} & 45.9459 & 1.8000 & 14.9000 & 71.7200 & 20.7700 & 42.1200 & 80.7400 & 0.2356 \\
\midrule
\multicolumn{18}{c}{\textit{Closed-Source Multimodal Models}} \\
\midrule
Doubao-1.5-Vision-Pro-32k   & 0.5580 & 0.2597 & 0.2922 & 0.2147 & 0.1700 & 0.0729 & 0.1240 & 0.3664 & 0.3377 & 33.1731 & 0.7100 & 6.6450 & 72.4000 & 8.6400 & 33.3000 & 78.4200 & 0.2587 \\
GPT-4o-Mini                 & 0.4924 & 0.3784 & 0.1922 & 0.1272 & 0.1357 & 0.0846 & 0.0960 & 0.2267 & 0.1976 & 19.2308 & 4.9400 & 17.5200 & 74.1300 & 11.7300 & 36.2900 & 77.5300 & 0.2388 \\
GPT-4o                      & 0.4928 & 0.4132 & 0.1504 & 0.0974 & 0.1161 & 0.0850 & 0.0840 & 0.3712 & 0.3527 & 15.7895 & 2.6800 & 14.7700 & 73.3500 & 33.7700 & 49.9600 & 81.5800 & 0.2253 \\
Gemini-1.5-Pro              & 0.3781 & 0.2247 & 0.0909 & 0.0476 & 0.2700 & 0.0661 & 0.0980 & 0.2772 & 0.2205 & 40.7051 & 0.5800 & 9.9400 & 70.5500 & 28.5800 & 45.9200 & 80.0200 & 0.1999 \\
Gemini-2.0-Pro-Exp          & 0.4925 & 0.4194 & 0.1648 & 0.1323 & 0.1714 & 0.0945 & 0.0820 & 0.1945 & 0.1498 & \resultthird{53.3333} & 0.2600 & 6.9200 & 40.2400 & 31.1800 & 48.6000 & 81.6000 & 0.2438 \\
Gemini-2.5-Pro-Preview      & 0.4256 & 0.3112 & 0.2098 & 0.1493 & 0.2709 & 0.2714 & \resultthird{0.2518} & 0.2937 & 0.2672 & 34.4970 & 5.5030 & 18.0180 & 74.4930 & 15.0110 & 38.0070 & 75.9890 & 0.2968 \\
Claude-3.7-Sonnet           & 0.2121 & 0.0449 & 0.1453 & 0.0479 & 0.1356 & 0.0540 & 0.0760 & 0.1764 & 0.1500 & 36.0215 & 0.6900 & 12.2300 & 68.7400 & 1.2900 & 16.6600 & 71.6600 & 0.1596 \\
Qwen-Max                    & 0.4566 & 0.2676 & 0.1925 & 0.0871 & 0.1606 & 0.0761 & 0.0940 & \resultone{0.1248} & \resultone{0.0843} & \resultone{69.2308} & 3.5000 & 17.0200 & 73.9600 & 30.6700 & 49.0000 & 82.5500 & 0.2445 \\
MedGemma-4B-it & 0.5005 & 0.4336 & 0.3071 & 0.1520 & 0.2750 & 0.0858 & 0.0200 & \resulttwo{0.1667} & \resulttwo{0.1316} & \resulttwo{55.0962} & 1.5360 & 15.0348 & 74.0205 & 4.8777 & 35.9803 & 76.7859 & 0.2668 \\
USFM-ViT & \resulttwo{0.6815} & \resultthird{0.5204} & 0.4791 & 0.3713 & - & - & - & - & - & - & - & - & - & - & - & - & 0.2321 \\
Dolphin V1-7B & \resultthird{0.6704} & \resultone{0.5743} & \resultthird{0.6681} & \resultthird{0.5329} & \resultthird{0.3185} & \resultthird{0.4380} & 0.2200 & 0.3131 & 0.2304 & 30.4534 & 2.7357 & 14.8181 & 73.0423 & \resultthird{41.4477} & \resultthird{58.5160} & \resultthird{89.3257} & \resultthird{0.4905} \\
Dolphin V1 & 0.6530 & \resulttwo{0.5667} & \resulttwo{0.6900} & \resulttwo{0.5791} & \resulttwo{0.4028} & \resulttwo{0.5302} & \resulttwo{0.3620} & 0.2717 & 0.2454 & 39.0725 & 3.8512 & 16.4303 & \resultthird{74.1854} & \resulttwo{49.0289} & \resulttwo{64.2466} & \resulttwo{91.1290} & \resulttwo{0.5390} \\
Dolphin R1 & \resultone{0.6819} & 0.5155 & \resultone{0.6943} & \resultone{0.5821} & \resultone{0.4775} & \resultone{0.6003} & \resultone{0.5080} & 0.2430 & 0.2273 & 38.6458 & 3.2193 & 15.1170 & 72.7287 & \resultone{54.0634} & \resultone{76.0111} & \resultone{92.9601} & \resultone{0.5835} \\
\bottomrule
\end{tabular}

}
\end{table}

\begin{figure*}[t]
\centering

\includegraphics[width= 1\linewidth]{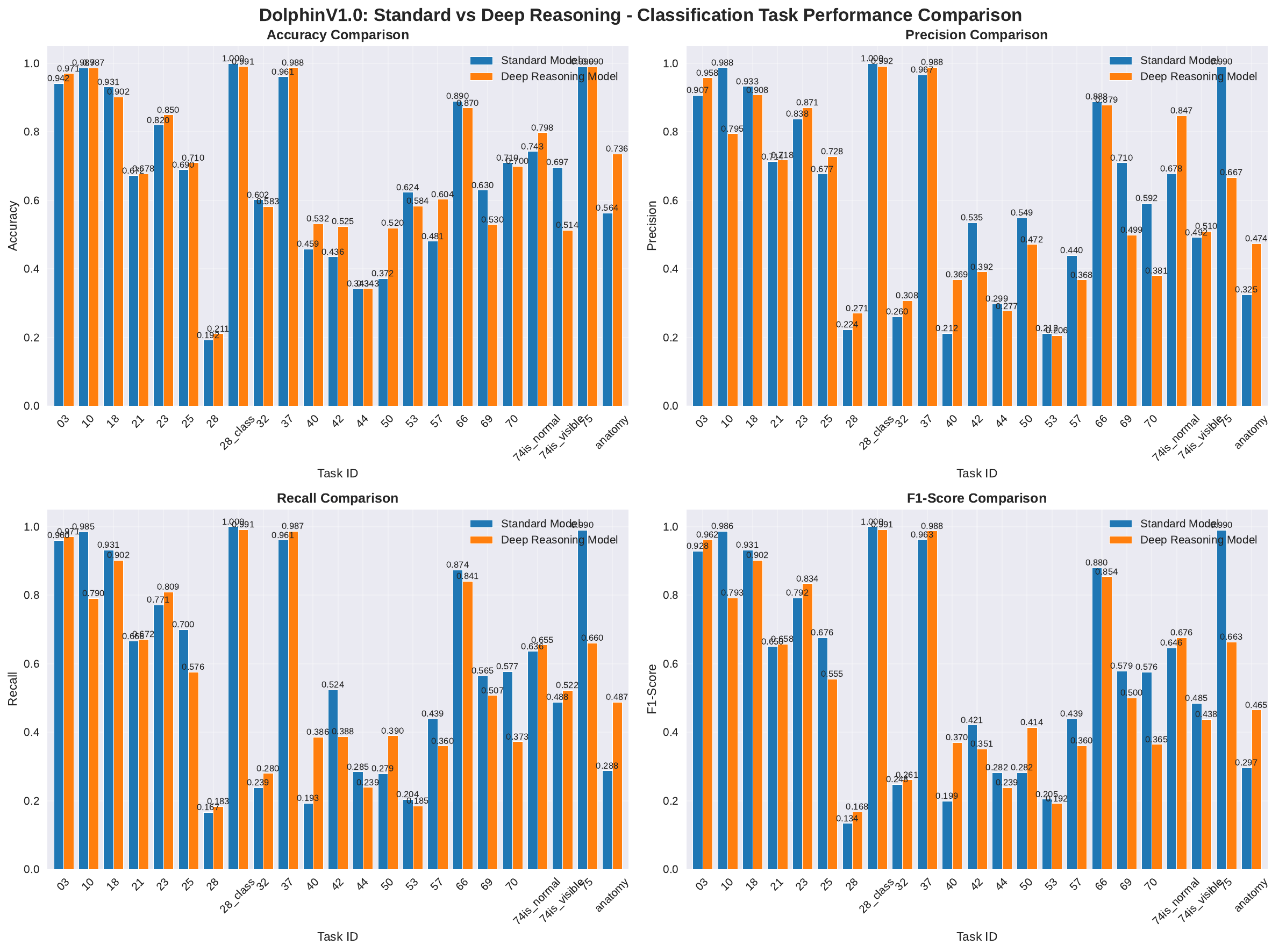}
\caption{Comparison of Standard and Deep Reasoning models on DolphinV1 classification tasks, evaluated by accuracy, precision, recall, and F1-score.}
\label{deep_reasoning}
\end{figure*}
\subsection{Performance Overview}
Dolphin R1 excels in classification and detection tasks, particularly in View Recognition and Assessment (VRA), where it achieves the highest accuracy (0.6943) and F1 score (0.5821), indicating strong spatial understanding and anatomical structure recognition capabilities. Similarly, it outperforms other models in Lesion Localization (LL), Organ Detection (OD), and Keypoint Detection (KD), with accuracy scores of 0.4775, 0.6003, and 0.5080 respectively, suggesting robust spatial reasoning and fine-grained visual recognition. These results highlight the model's ability to handle complex ultrasound-specific visual patterns and anatomical variations. Despite strong performance in classification and detection tasks, Dolphin R1 exhibits limitations in Clinical Value Estimation (CVE) and Report Generation (RG), achieving only moderate accuracy in numerical parameter.

In comparison, Dolphin V1 shows slightly lower but still strong performance across these tasks, while Dolphin V1-7B, as a smaller variant, exhibits a noticeable performance drop, particularly in OD and KD, indicating the importance of model scale in capturing fine-grained visual features.

\subsection{Impact of Model Scale}
The comparison between Dolphin V1-7B and its larger 72B counterparts—Dolphin V1 and Dolphin R1—highlights the significant impact of model scale on multimodal understanding in ultrasound-related tasks. The 72B models consistently outperform the smaller 7B variant across classification, detection, and generation tasks, with the most notable improvements seen in Organ Detection (OD), Keypoint Detection (KD), and Caption Generation (CG), indicating that larger models are better suited to capture the visual complexity and semantic nuances present in medical ultrasound imaging.

Moreover, Dolphin R1 achieves a significantly improved U2-Score compared to Dolphin V1, demonstrating that advanced training strategies such as GRPO can effectively enhance model performance. This highlights the importance of both model scale and training methodology in pushing the boundaries of ultrasound image understanding.

\subsection{Comparison Between Standard Mode and Deep Reasoning Mode}

As shown in Figure \ref{deep_reasoning}, we also evaluated the performance of our model, Dolphin R1, on Disease Diagnosis (DD) tasks under both standard mode and deep reasoning mode. The results demonstrate that, in the majority of tasks, enabling the deep reasoning mode leads to a notable improvement in diagnostic accuracy. Moreover, the deep reasoning mode offers a more thorough and detailed analytical process, providing interpretable and transparent reasoning trails. This enhanced interpretability not only supports more informed clinical decision-making but also aligns better with user preferences, as evidenced by the fact that physicians tend to favor the deep reasoning mode in practice.

\section{Conclusion}
In this work, we introduce Dolphin V1 and Dolphin R1, advancing multimodal ultrasound understanding through large-scale data curation and progressive training strategies. The 2-million-scale dataset provides the foundation for robust learning across heterogeneous tasks, while the integration of reinforcement learning in Dolphin R1 endows the model with superior reasoning and interpretability. Our experimental results show that Dolphin R1 achieved a U2-score of 0.5835, surpassing all competitors by a large margin.
The comparative analysis highlights that deep reasoning not only improves quantitative accuracy but also enhances the interpretability of diagnostic processes, aligning more closely with physician preferences. Looking forward, this dual-track design of standard vs. deep reasoning modes provides a flexible paradigm for future medical AI systems.

\section{Authors}
\textbf{Core Contributors:} Taohan Weng

\textbf{Contributors\footnote{Alphabetical order.}:} Xiaoqing Guo, Kaibing Hu, Henan Liu, Siya Liu, Xiaoyang Liu, Zhenyu Liu, Jiren Ren, 
Boyan Wang, Boyang Wang, Yiyu Wang, Yalun Wu, Chaoran Yan, Kaiwen Yan, 
Jinze Yu, Chi Zhang, Duo Zhang, Haoyun Zheng

\textbf{Project Leader:} Hongcheng Guo, Anjie Le

\textbf{Project Advisor:} Jacques Souquet

\section*{Acknowledgement}
We sincerely thank all the participating engineers and clinicians for their invaluable contributions throughout this project, including data labeling, validation, and quality checking. Their expertise and dedication ensured the clinical reliability of the datasets and the robustness of our evaluation.

We extend our special gratitude to \textbf{Esonic Image} for their generous support, including funding, provision of ultrasound devices for data collection, and coordination of medical experts to annotate datasets and provide continuous feedback during model development. Their collaboration was essential in bridging technical innovation with real-world clinical needs.

\newpage
\bibliography{colm2024_conference}
\bibliographystyle{colm2024_conference}

\newpage

\appendix
\section{Appendix}

\begin{figure*}[ht]
\centering

\includegraphics[width= 0.95\linewidth]{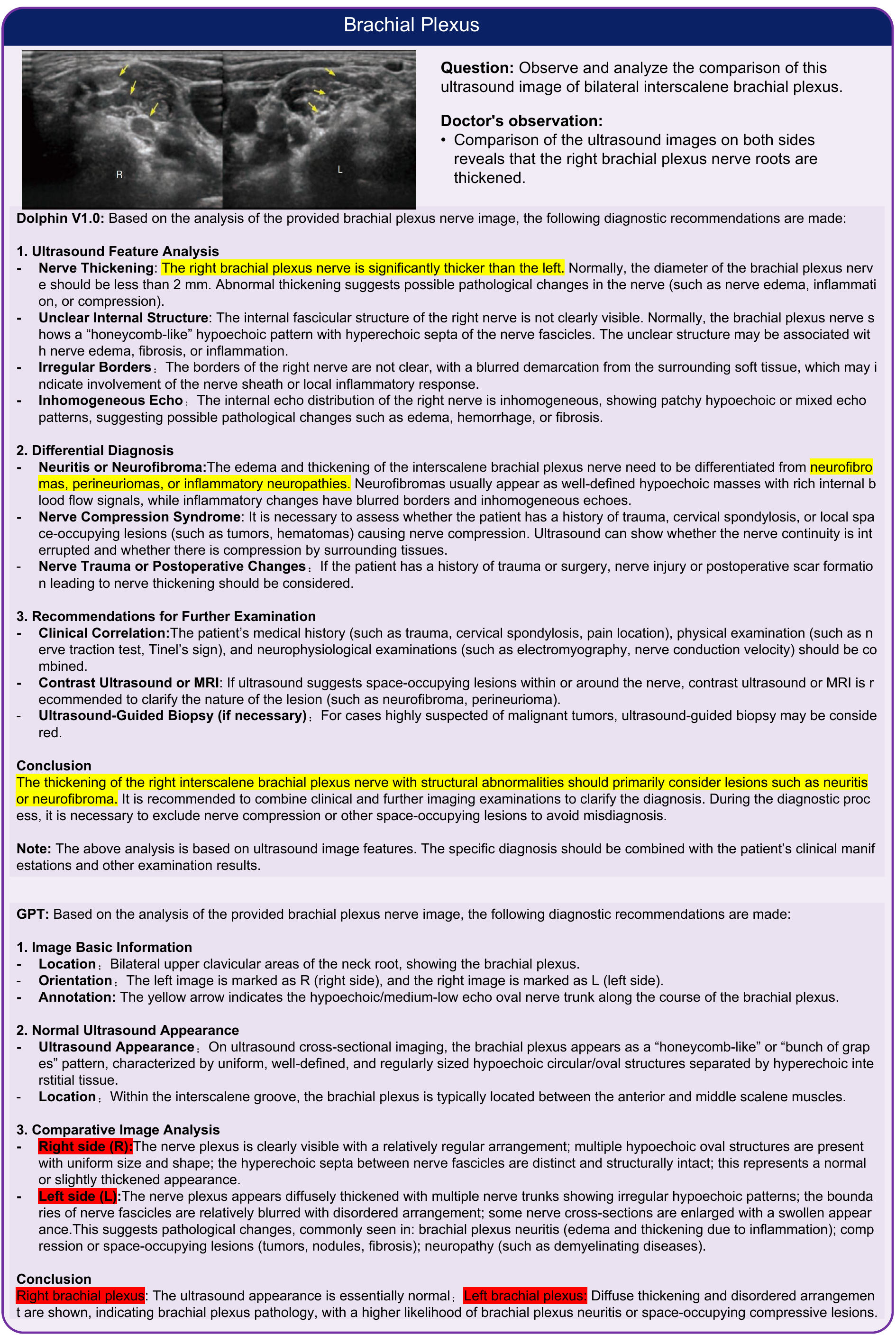}

\label{brachial_plexus_nerve_root_thickening}
\end{figure*}

\begin{figure*}[ht]
\centering

\includegraphics[width= 0.95\linewidth]{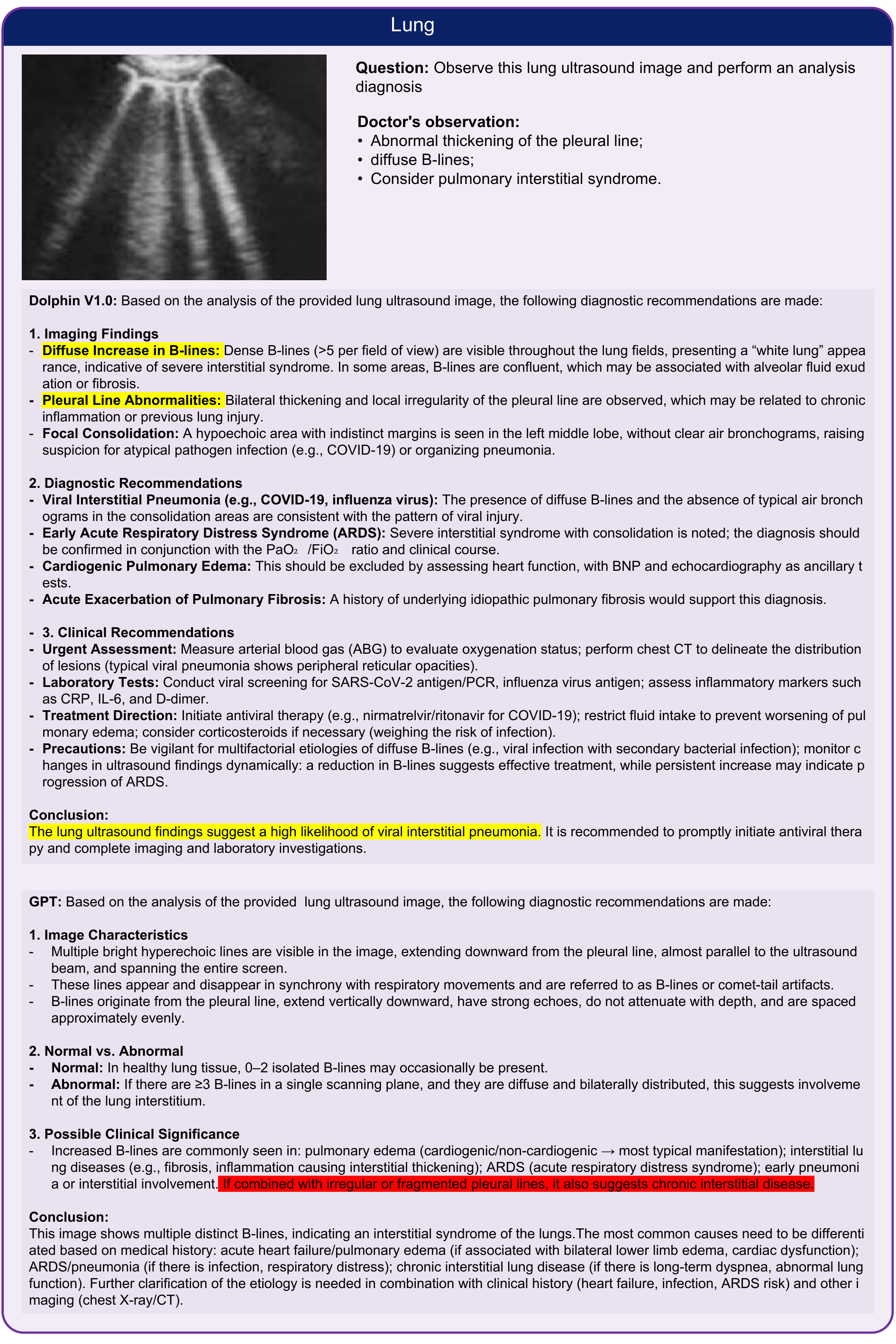}

\label{lung_b_lines}
\end{figure*}

\begin{figure*}[ht]
\centering

\includegraphics[width= 0.95\linewidth]{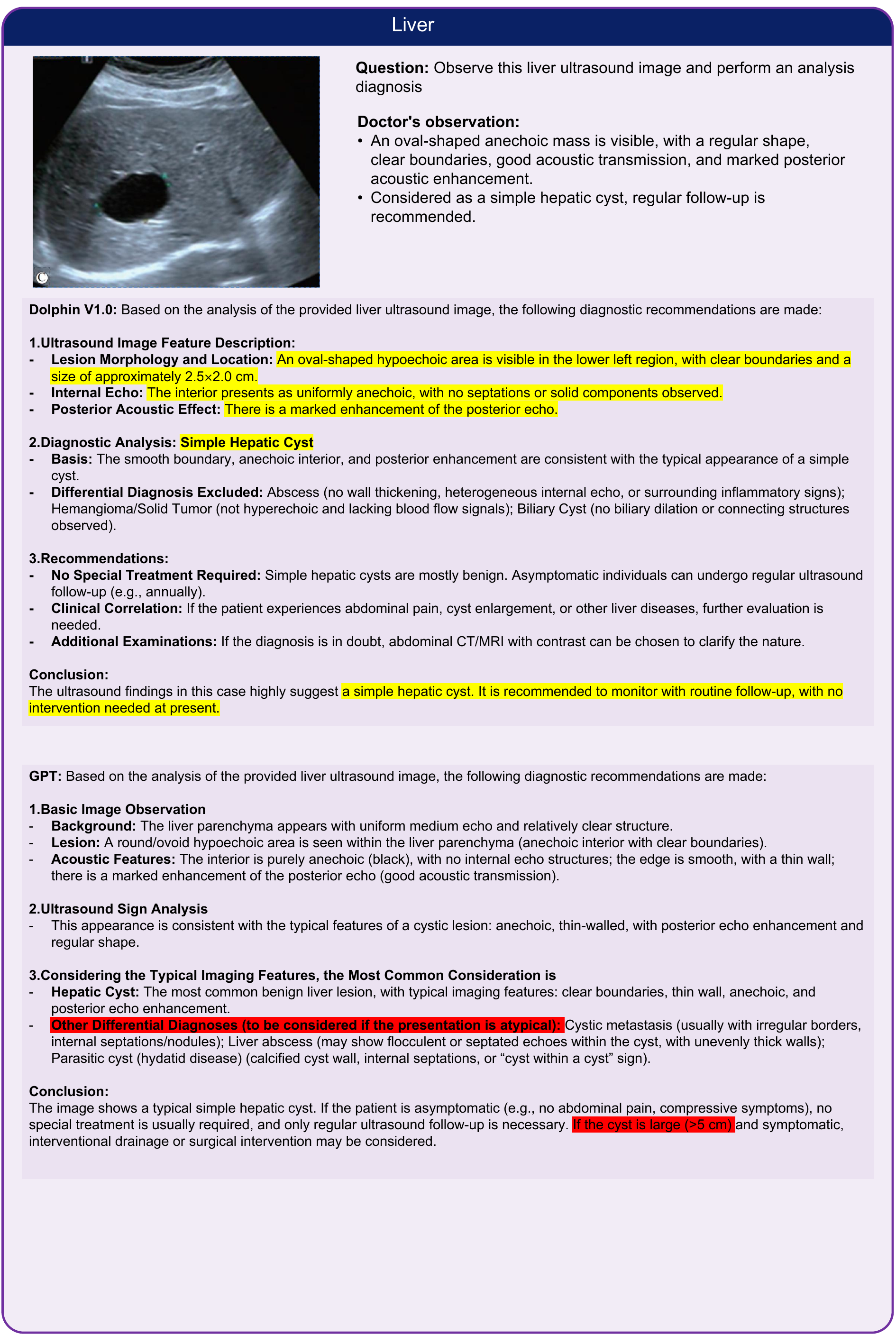}

\label{liver_hepatic_cyst}
\end{figure*}

\begin{figure*}[ht]
\centering

\includegraphics[width= 0.95\linewidth]{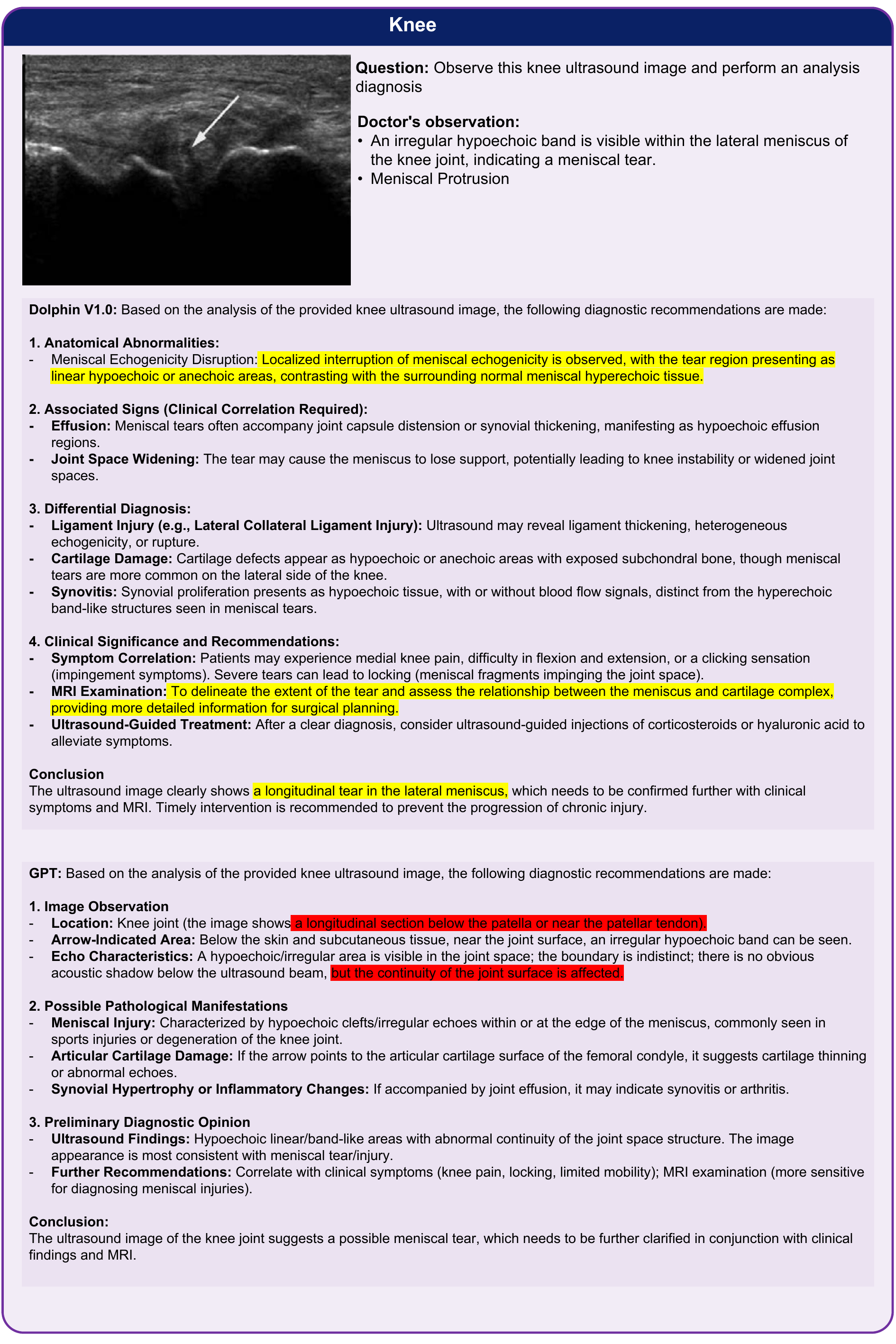}

\label{knee_lateral_meniscus_tear}
\end{figure*}

\begin{figure*}[ht]
\centering

\includegraphics[width= 0.95\linewidth]{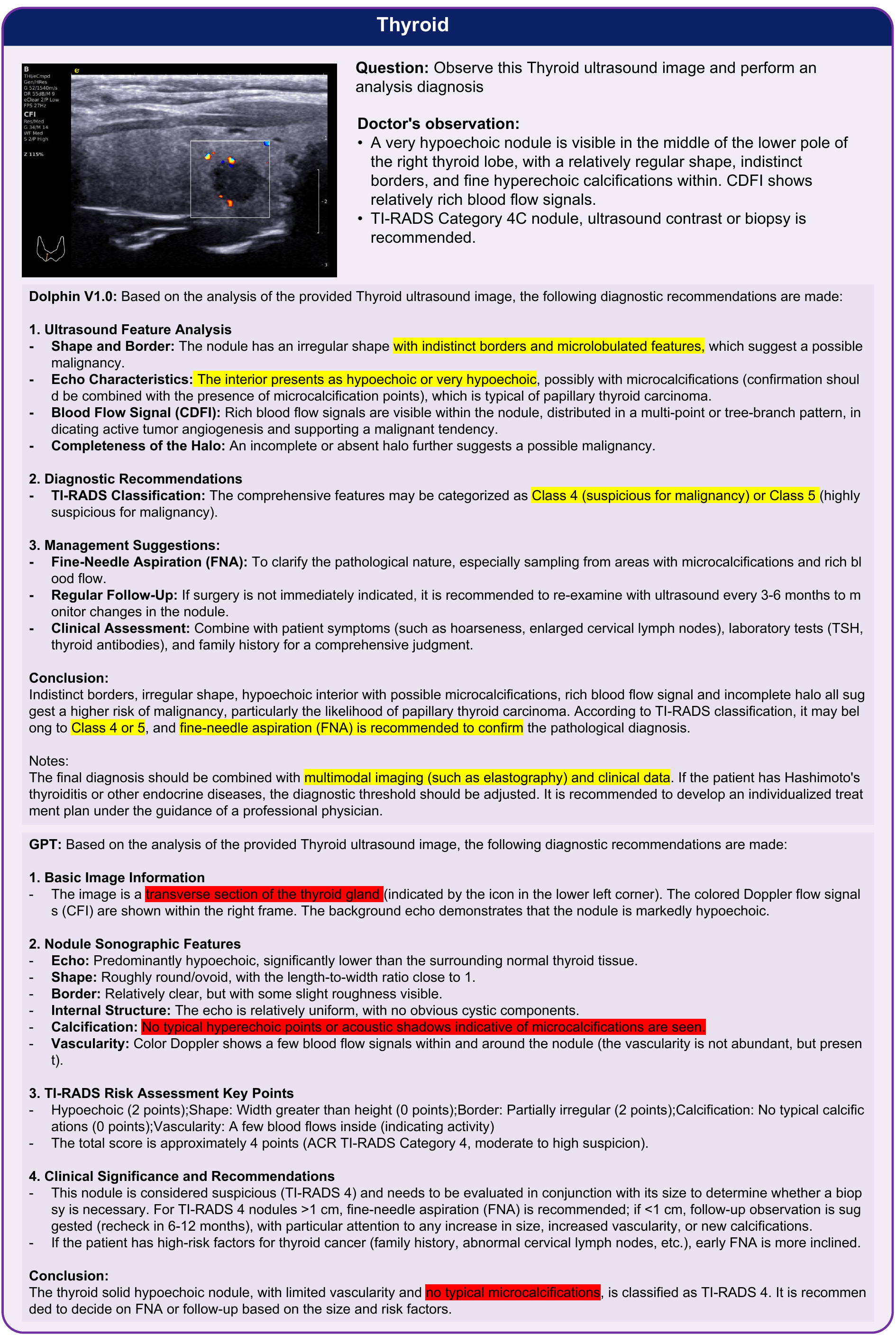}

\label{thyroid_malignant_nodule}
\end{figure*}

\begin{figure*}[ht]
\centering

\includegraphics[width= 0.95\linewidth]{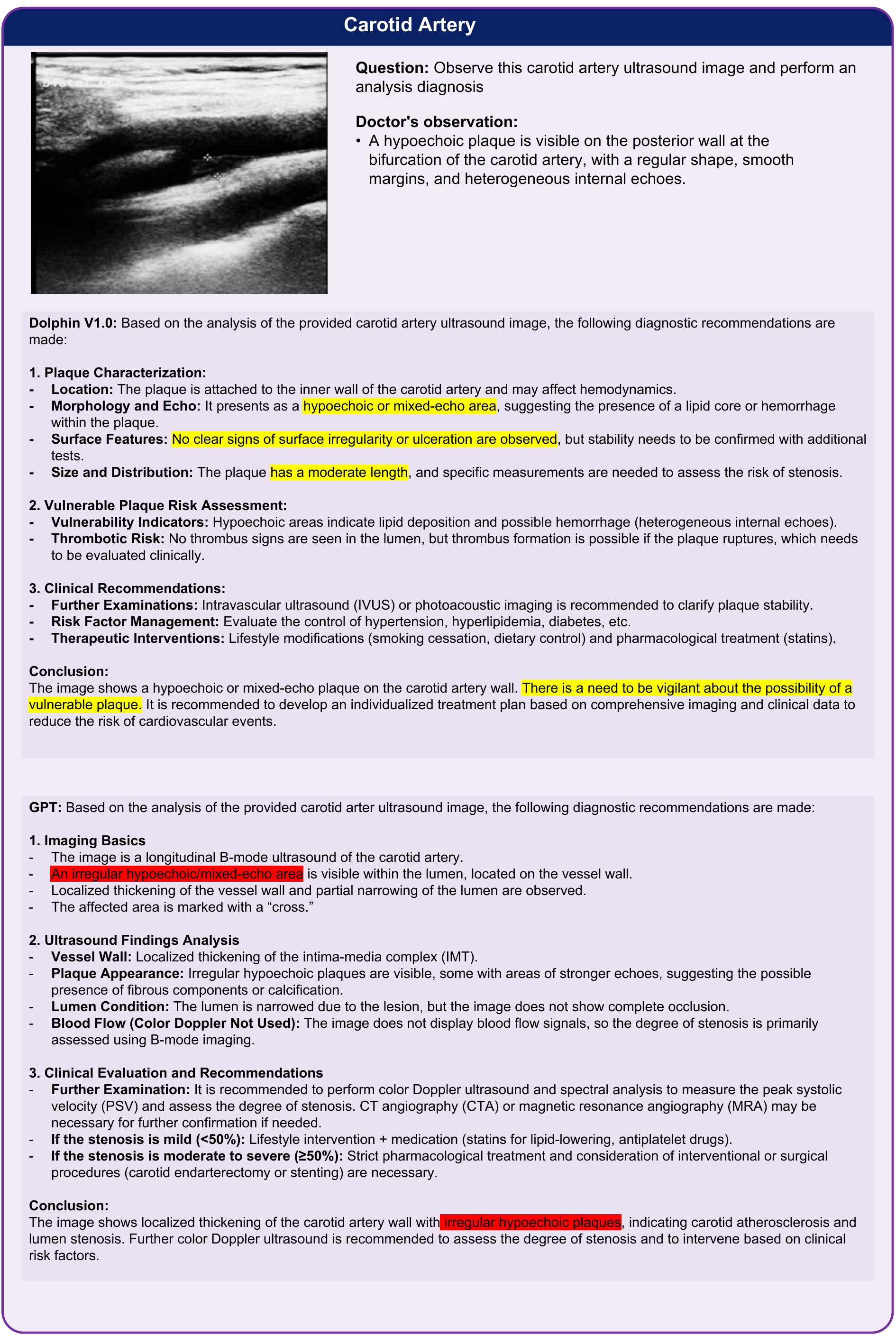}

\label{carotid_artery_plaque_posterior_wall}
\end{figure*}

\begin{figure*}[ht]
\centering

\includegraphics[width= 0.95\linewidth]{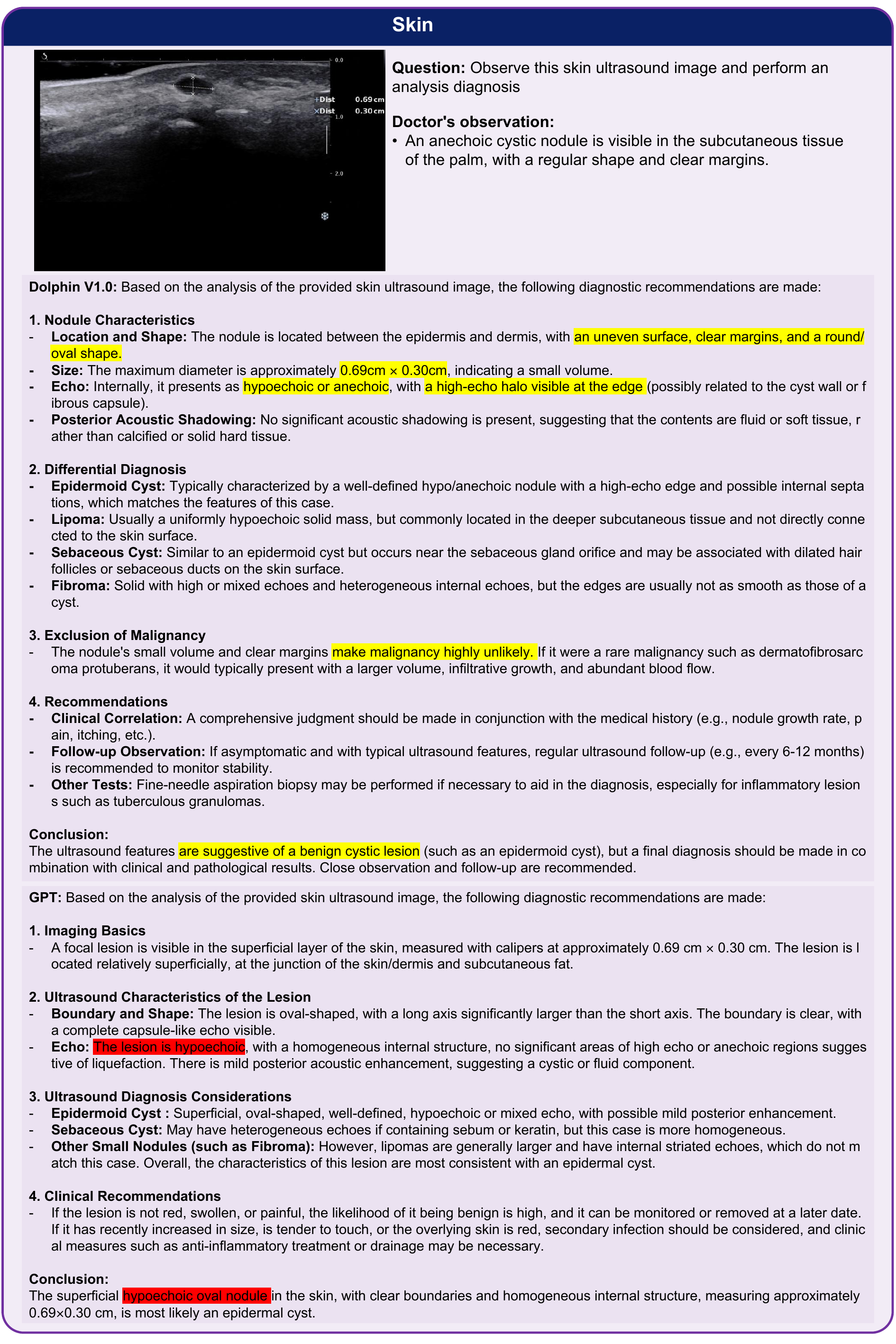}

\label{palm_subcutaneous_nodule}
\end{figure*}

\begin{figure*}[ht]
\centering

\includegraphics[width= 0.95\linewidth]{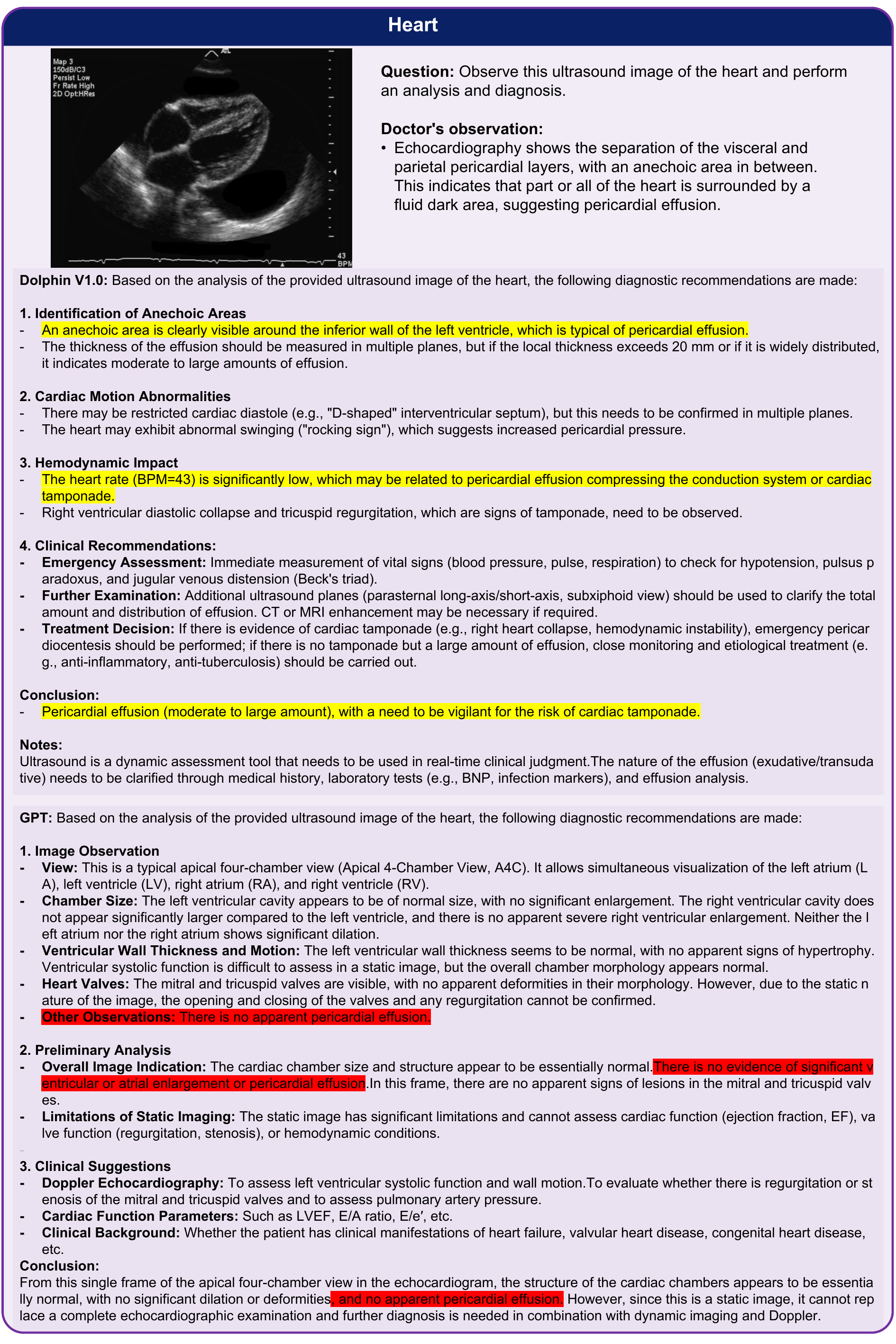}

\label{heart_pericardial_effusion}
\end{figure*}

\begin{figure*}[ht]
\centering

\includegraphics[width= 0.95\linewidth]{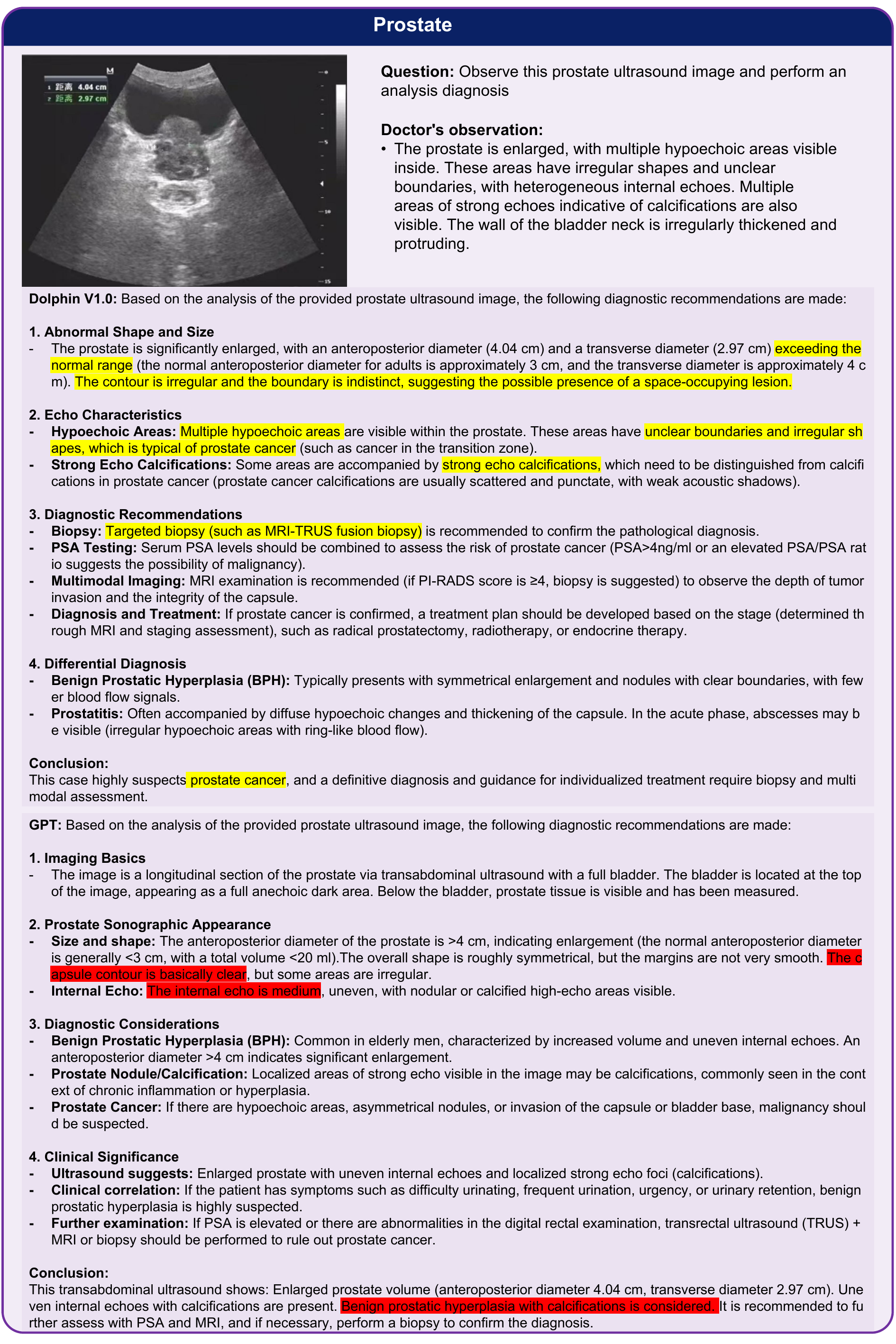}

\label{prostate_cancer}
\end{figure*}

\begin{figure*}[ht]
\centering

\includegraphics[width= 0.95\linewidth]{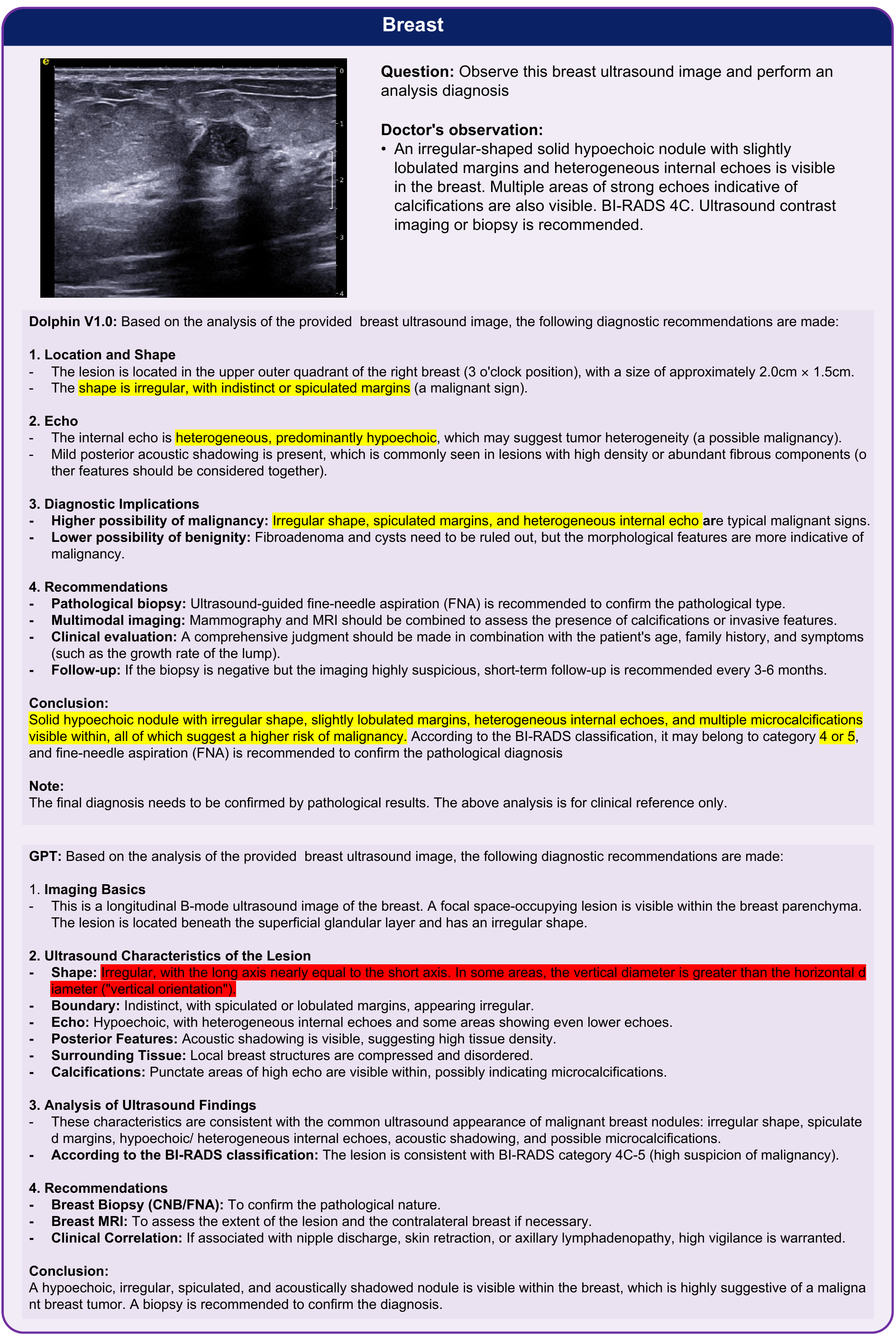}

\label{breast_malignant_nodule}
\end{figure*}

\begin{figure*}[ht]
\centering

\includegraphics[width= 0.95\linewidth]{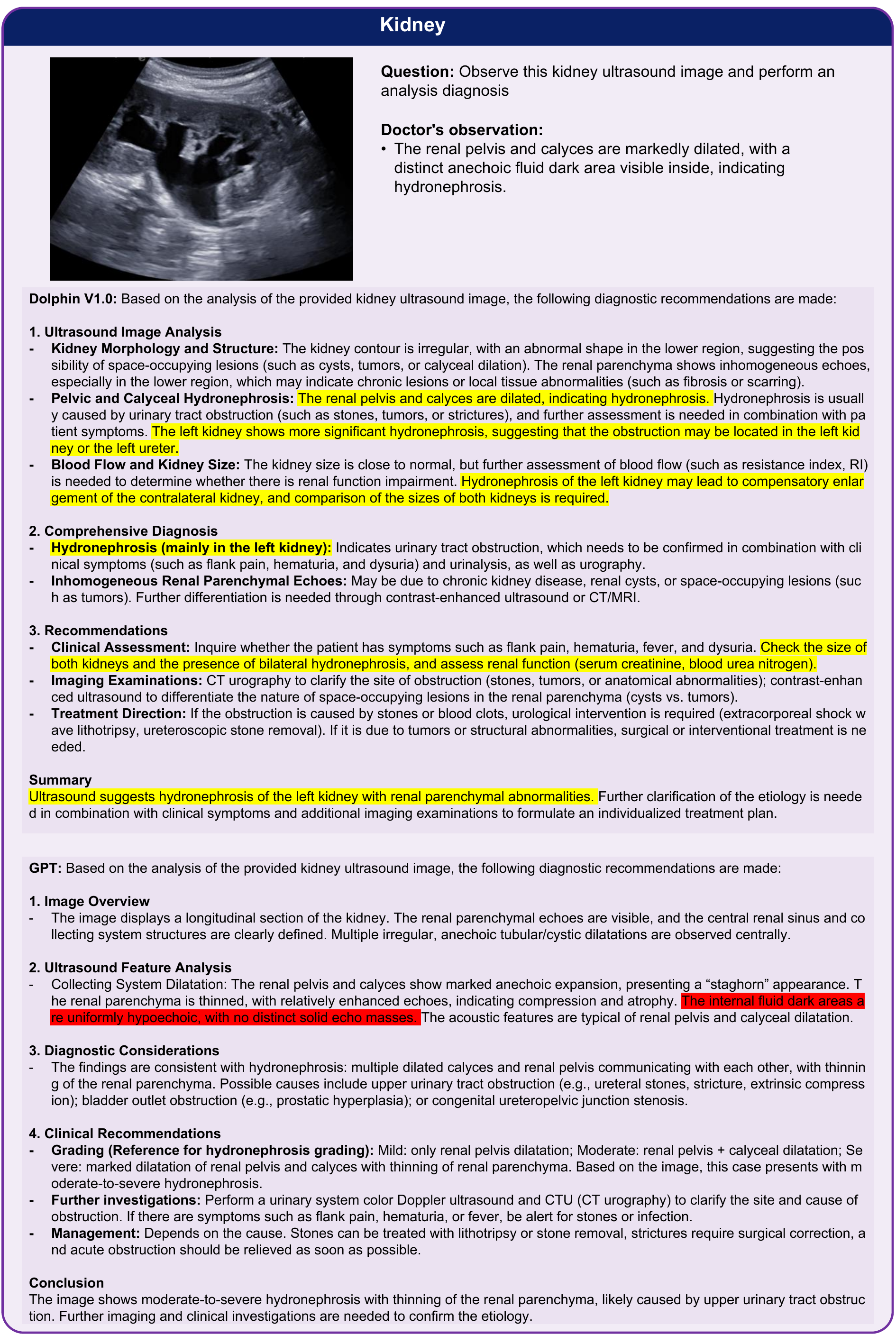}

\label{kidney_hydronephrosis}
\end{figure*}

\begin{figure*}[ht]
\centering

\includegraphics[width= 0.95\linewidth]{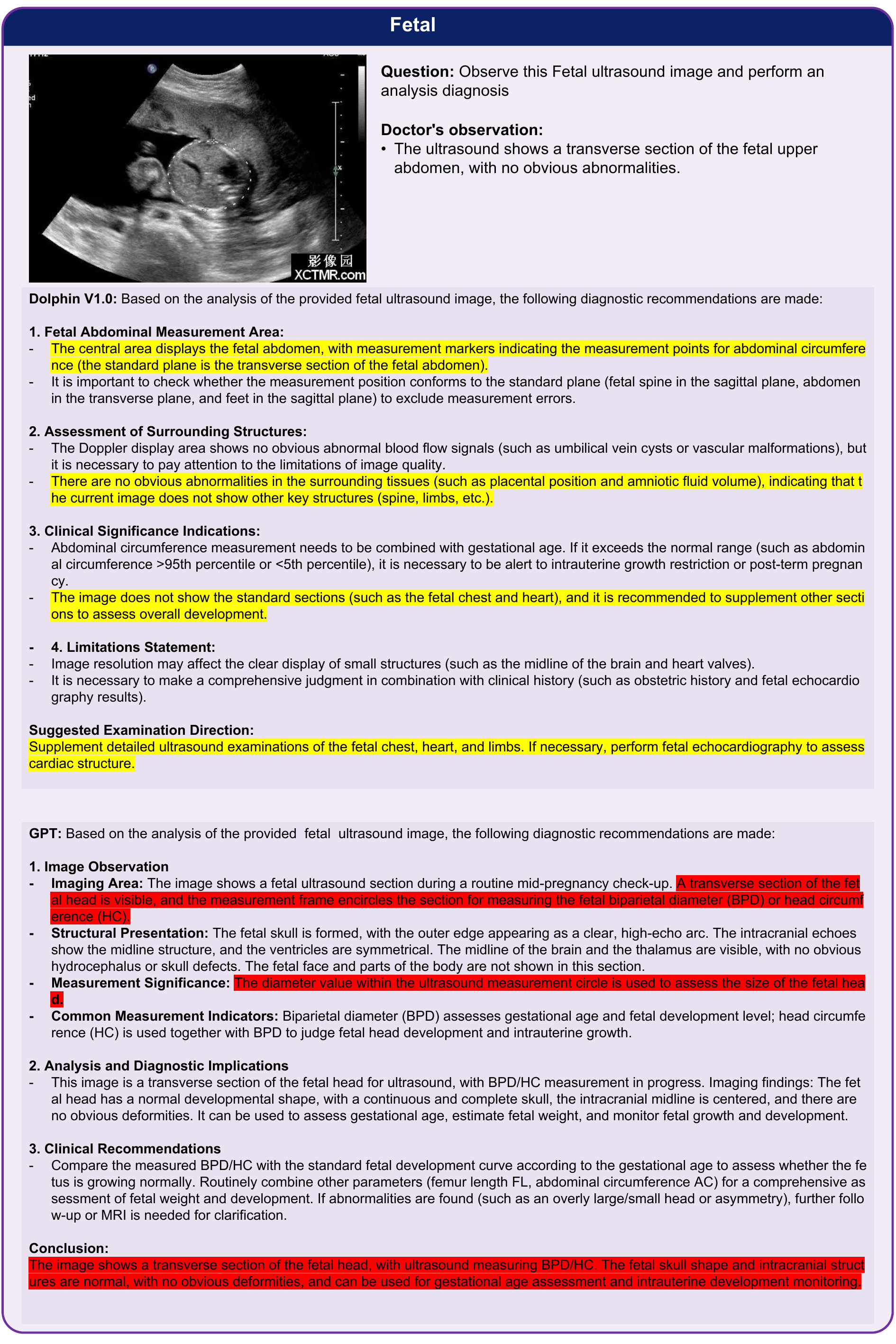}

\label{fetal_abdomen}
\end{figure*}

\begin{figure*}[ht]
\centering

\includegraphics[width= 0.95\linewidth]{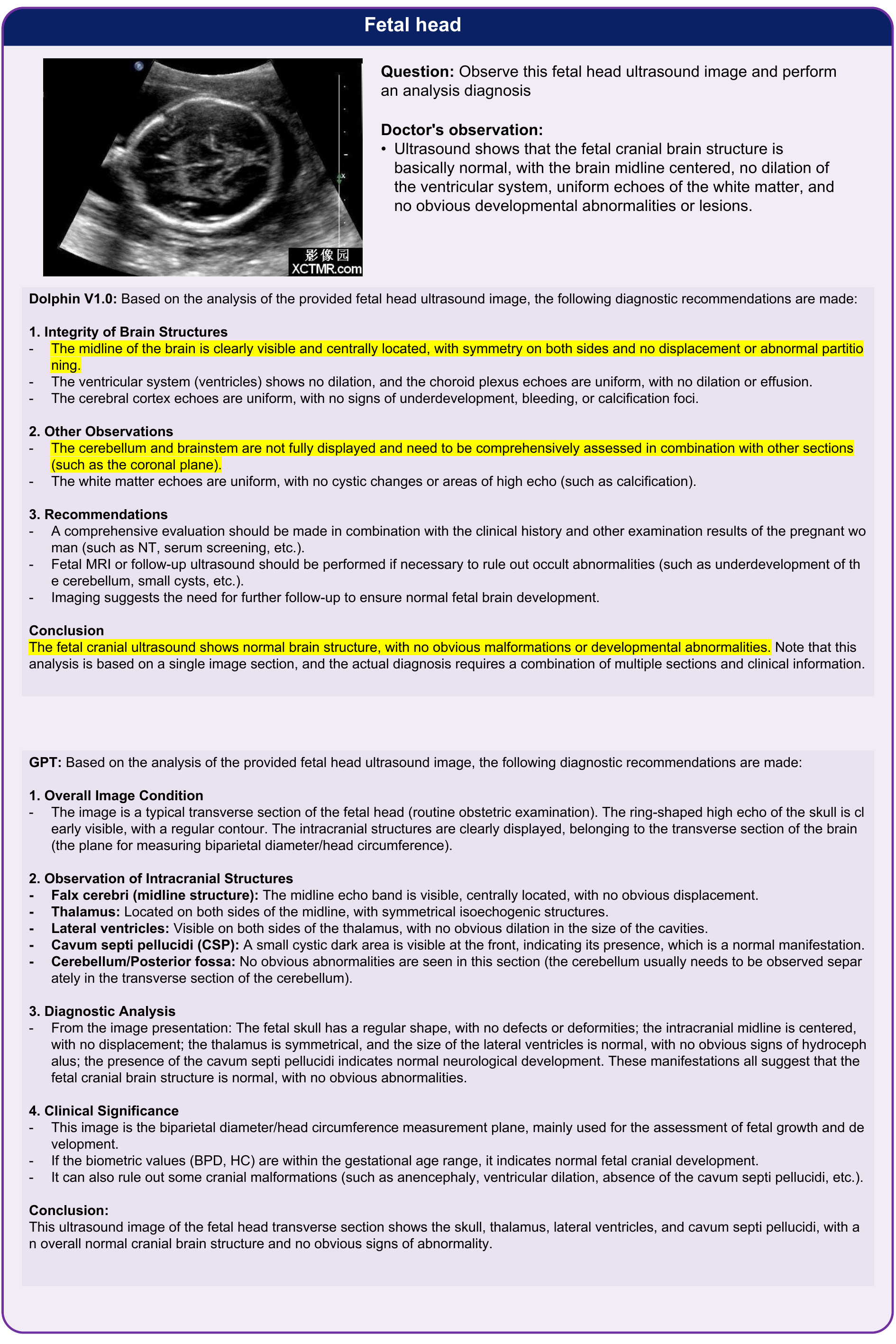}

\label{fetal_head}
\end{figure*}

\begin{figure*}[ht]
\centering

\includegraphics[width= 0.95\linewidth]{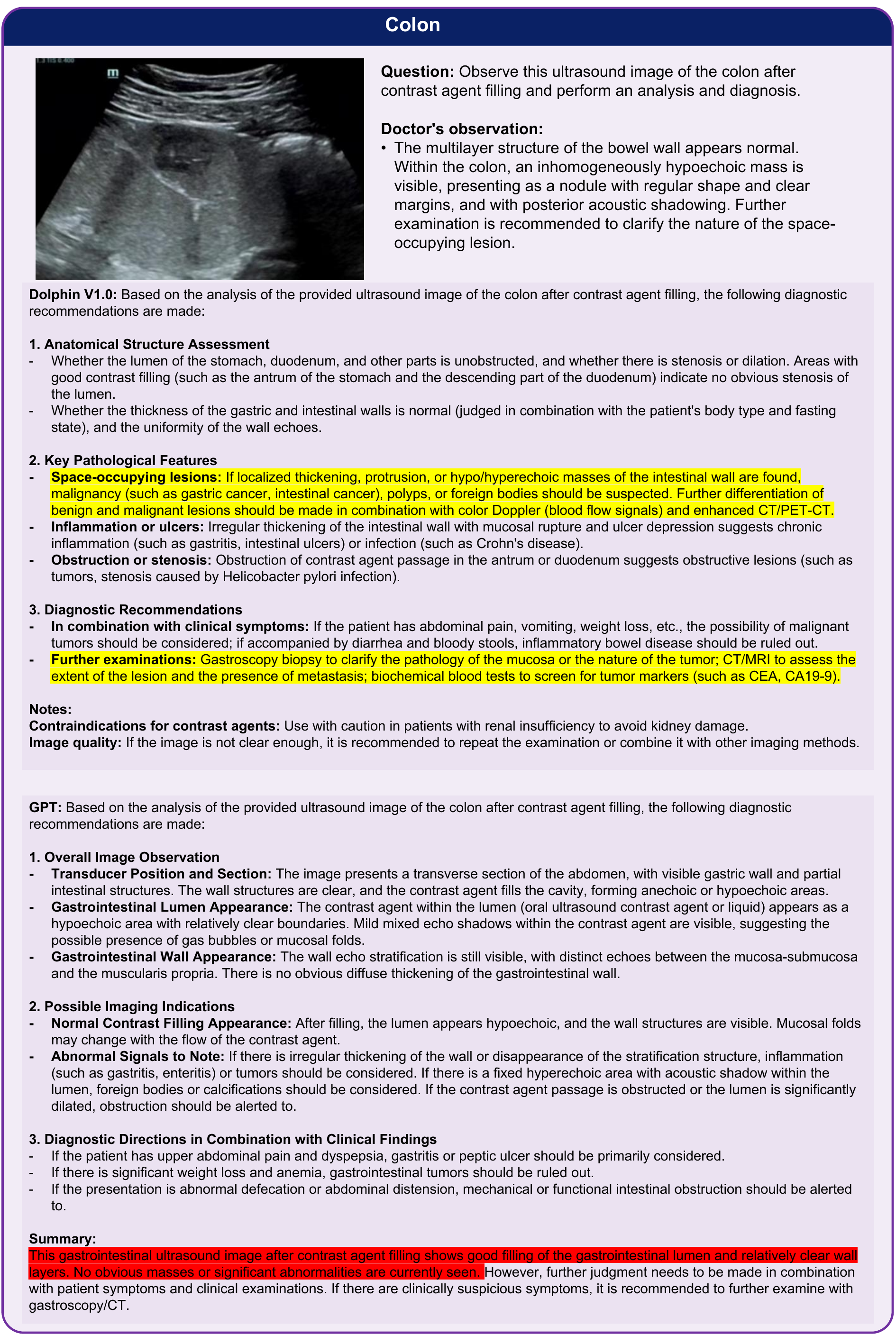}

\label{colon_polyp}
\end{figure*}

\begin{figure*}[ht]
\centering

\includegraphics[width= 0.95\linewidth]{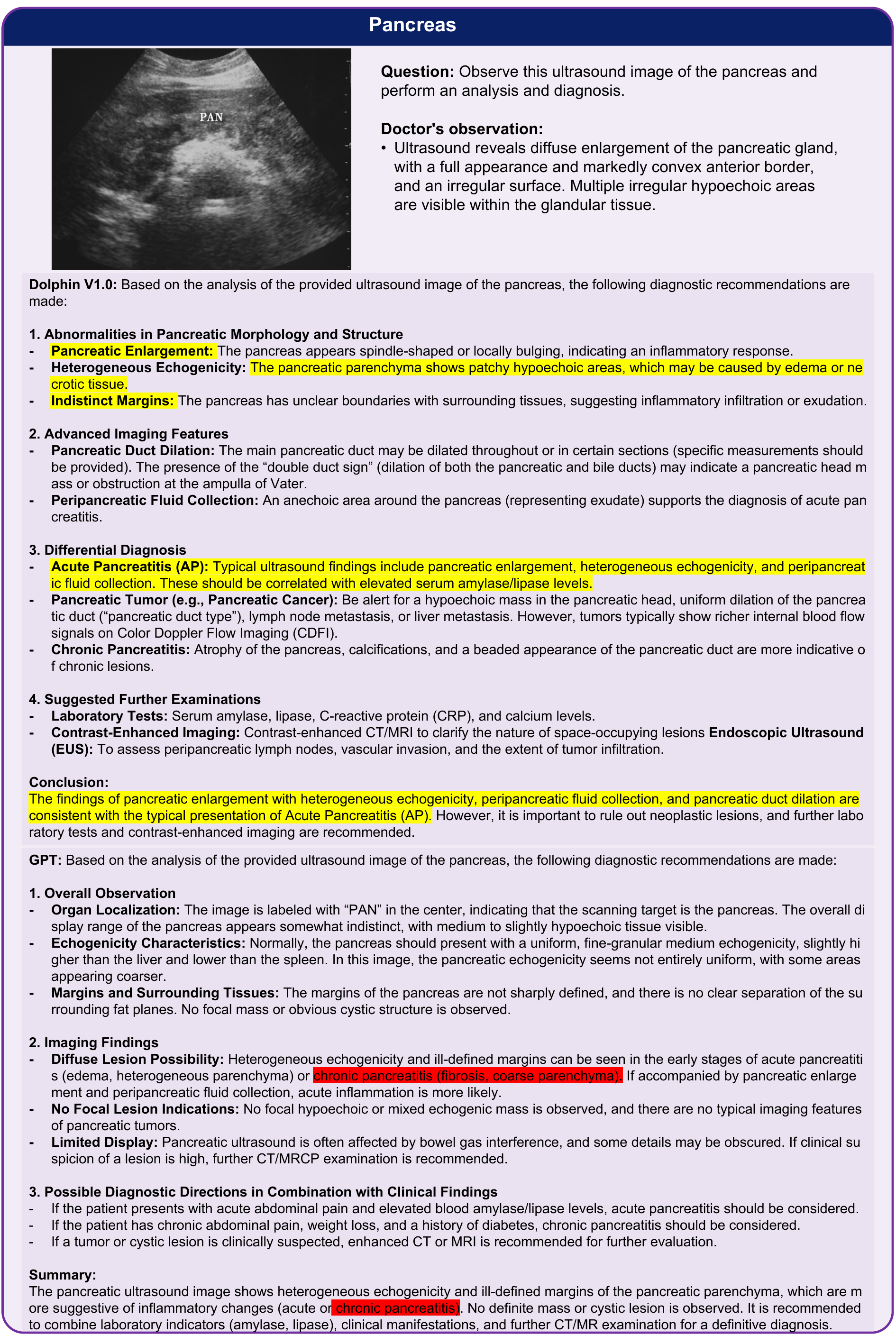}

\label{pancreas_acute_pancreatitis}
\end{figure*}

\begin{figure*}[ht]
\centering

\includegraphics[width= 0.95\linewidth]{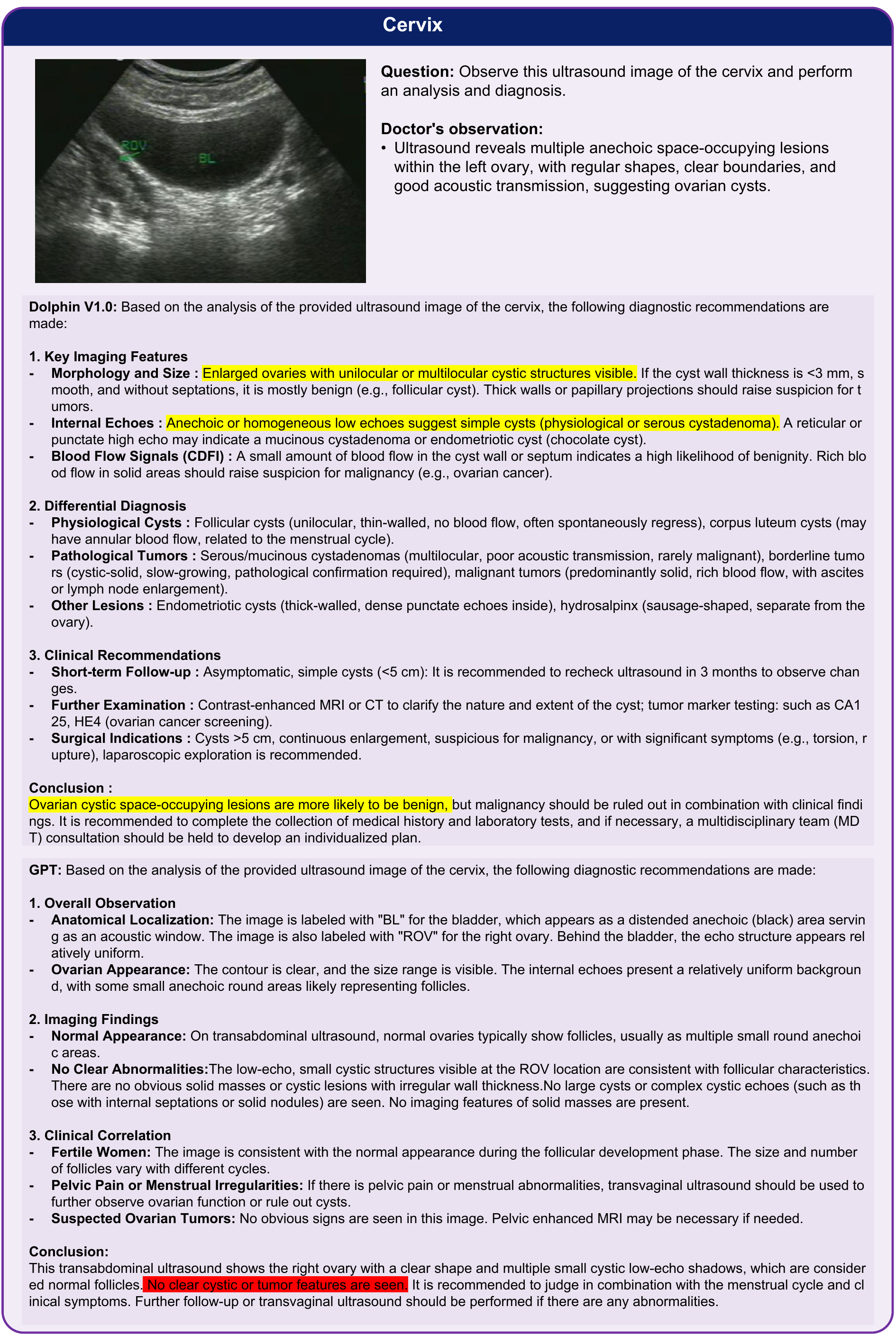}

\label{cervix_ovarian_cyst}
\end{figure*}

\begin{figure*}[ht]
\centering

\includegraphics[width= 0.95\linewidth]{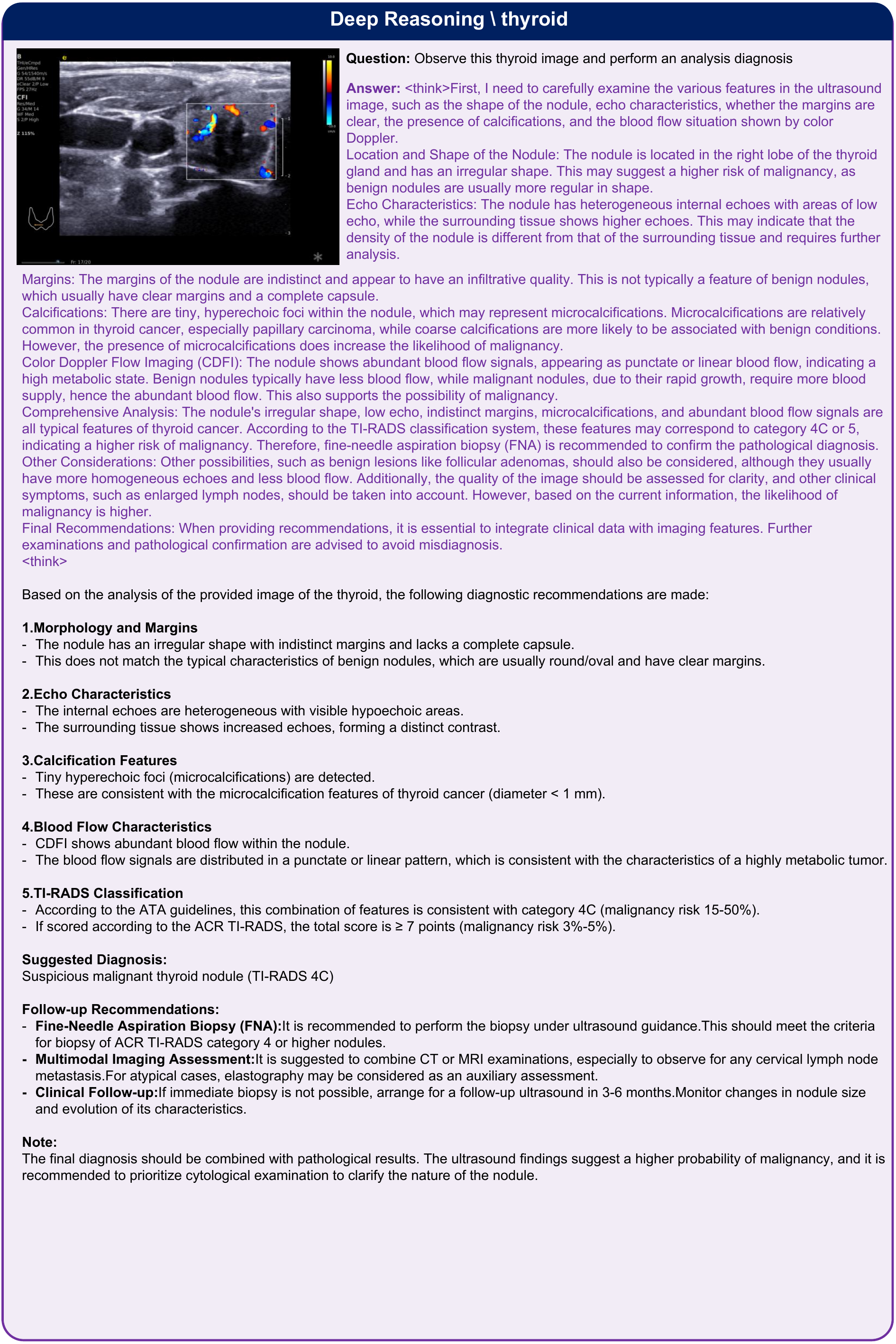}

\label{deep_reasoning_thyroid_malignant_nodule}
\end{figure*}

\begin{figure*}[ht]
\centering

\includegraphics[width= 0.95\linewidth]{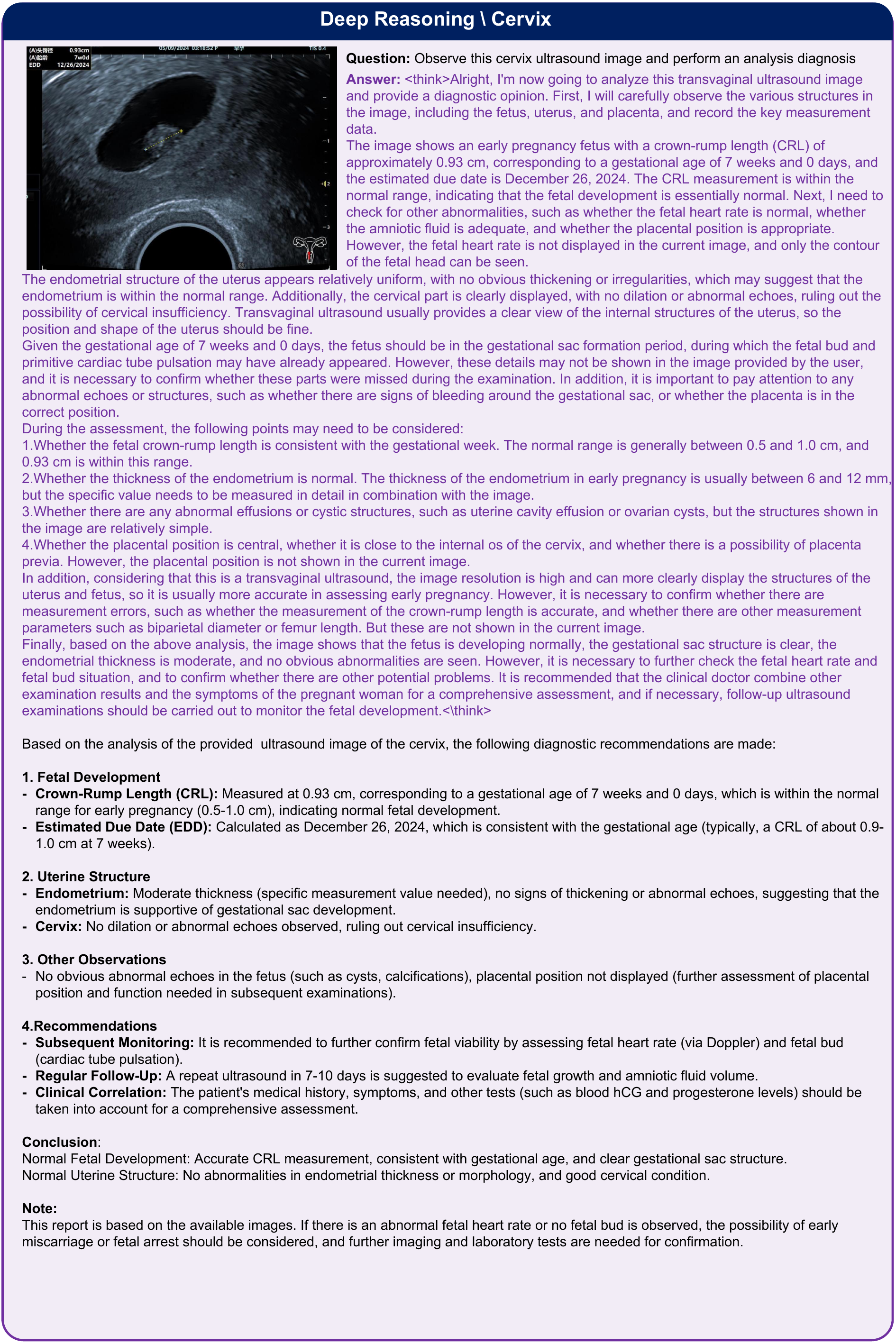}

\label{deep_reasoning_cervix}
\end{figure*}

\begin{figure*}[ht]
\centering

\includegraphics[width= 0.95\linewidth]{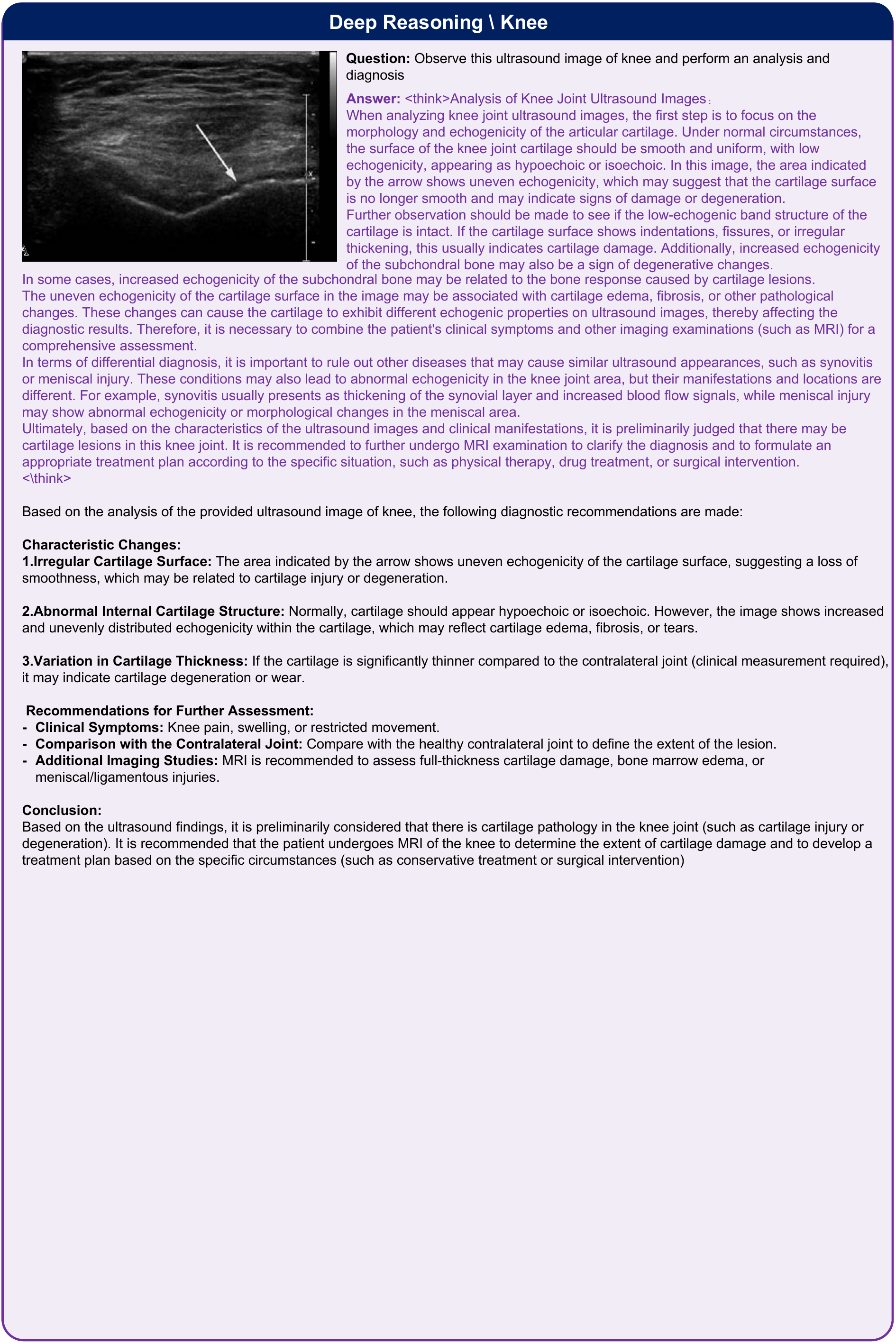}

\label{deep_reasoning_knee_cartilage_lesion}
\end{figure*}

\begin{figure*}[ht]
\centering

\includegraphics[width= 0.95\linewidth]{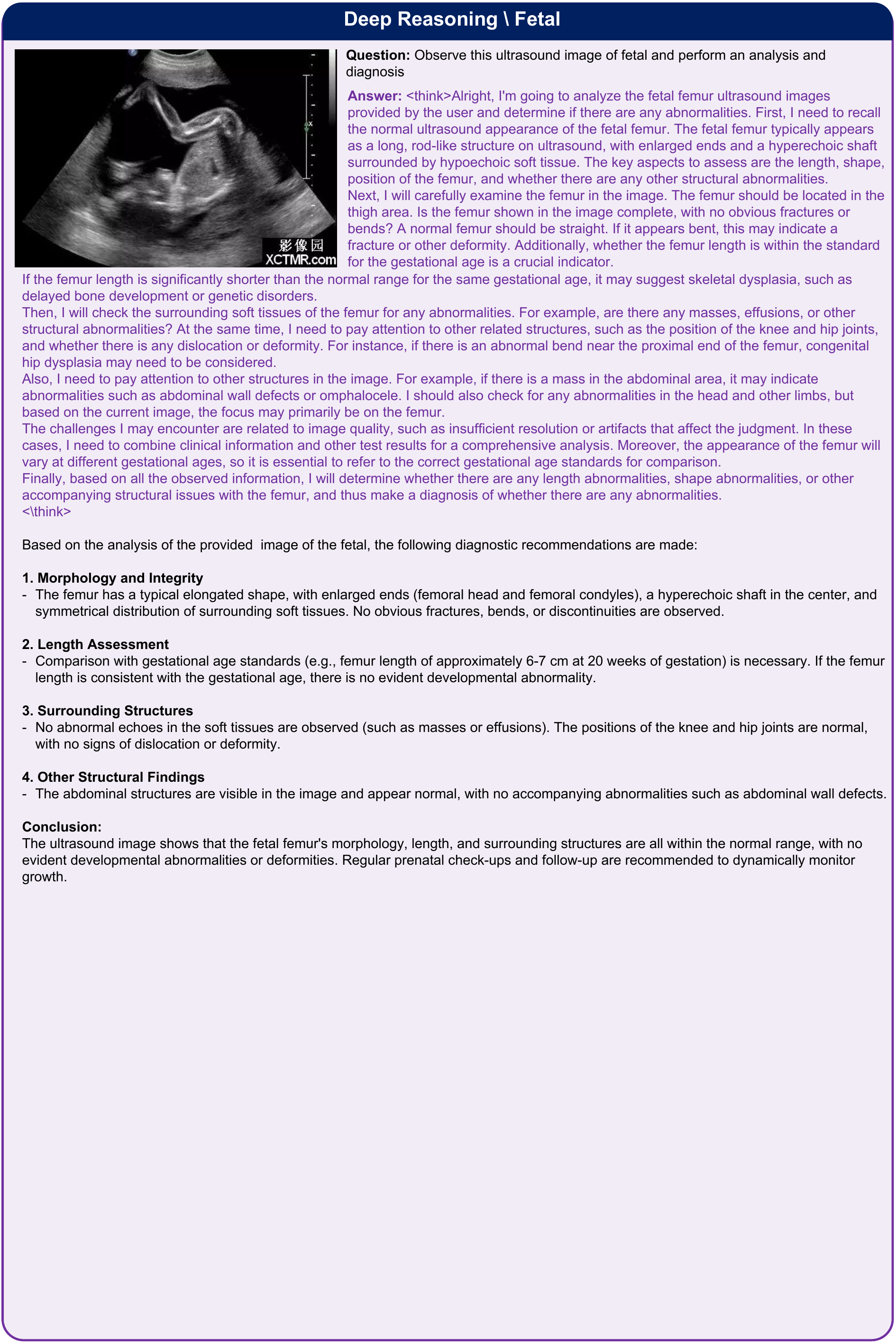}

\label{deep_reasoning_fetal_femur}
\end{figure*}

\begin{figure*}[ht]
\centering

\includegraphics[width= 0.95\linewidth]{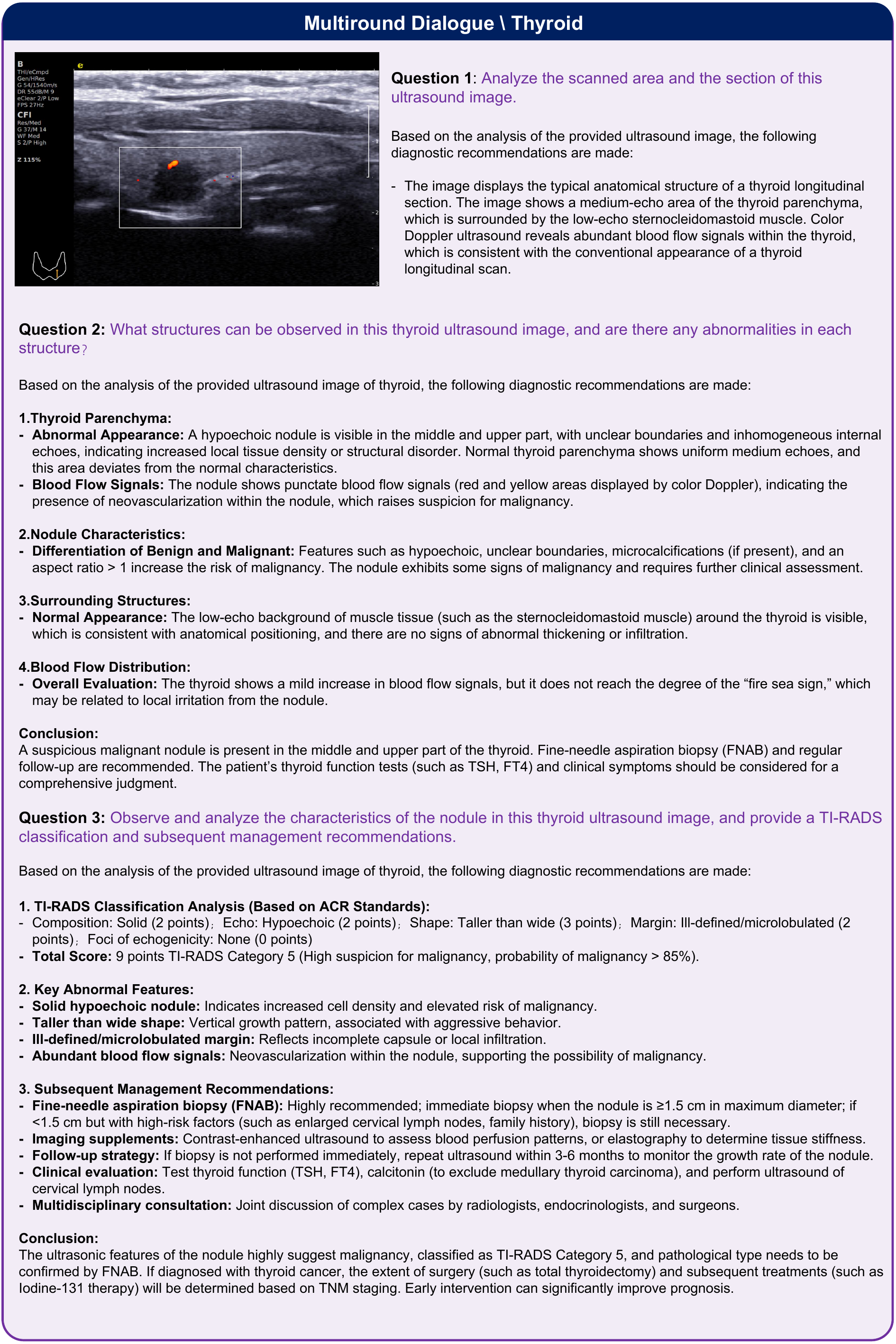}

\label{multi_turn_dialogue_thyroid}
\end{figure*}

\begin{figure*}[ht]
\centering

\includegraphics[width= 0.95\linewidth]{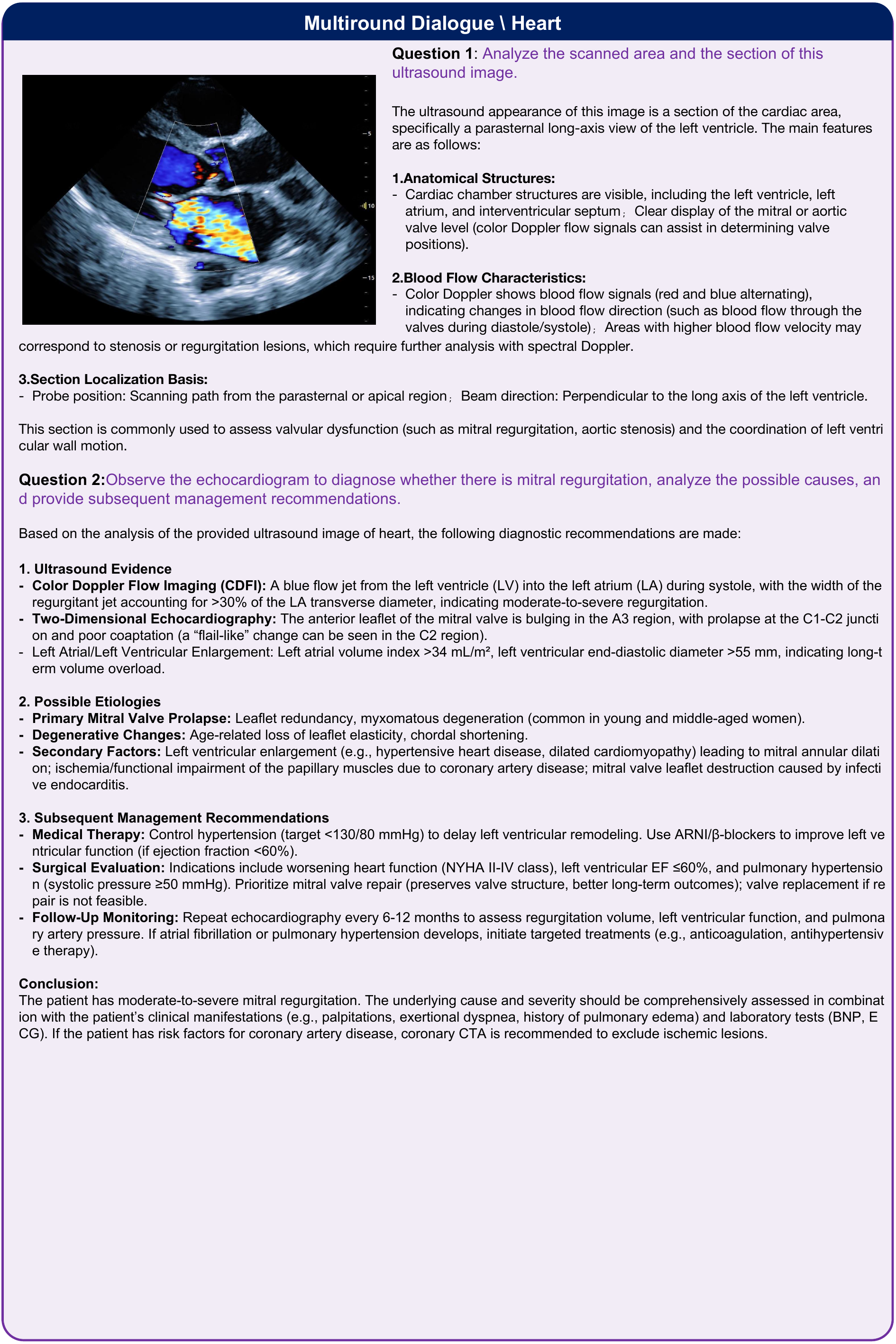}

\label{multi_turn_dialogue_heart}
\end{figure*}

\end{document}